\documentclass{article}
\usepackage{iclr2027_conference,times}

\usepackage{booktabs,multirow,graphicx,xcolor,colortbl}

\usepackage{fix-cm}
\usepackage{amsmath}
\usepackage{booktabs}
\usepackage{graphicx}
\usepackage[hidelinks]{hyperref}
\usepackage{microtype}
\usepackage{url}
\usepackage{xcolor}
\newcommand{\thinkopen}{\texttt{\textless think\textgreater}}
\newcommand{\thinkclose}{\texttt{\textless/think\textgreater}}
\newcommand{\answeropen}{\texttt{\textless answer\textgreater}}
\newcommand{\answerclose}{\texttt{\textless/answer\textgreater}}
\usepackage{booktabs}
\usepackage{multirow}
\usepackage{graphicx}
\usepackage{amssymb}
\usepackage{subcaption}
\usepackage{dblfloatfix}
\usepackage{amsmath,amsfonts,bm}

\def\1{\bm{1}}

\def\rvx{{\mathbf{x}}}

\DeclareMathAlphabet{\mathsfit}{\encodingdefault}{\sfdefault}{m}{sl}
\SetMathAlphabet{\mathsfit}{bold}{\encodingdefault}{\sfdefault}{bx}{n}

\newcommand{\cmark}{\checkmark}
\newcommand{\xmark}{--}

\title{\textsc{Merit}: Multimodal Evidence Reasoning over Image Tables}

\author{%
\makebox[\dimexpr\textwidth-2\tabcolsep\relax][c]{%
\fontsize{11pt}{13pt}\selectfont
\begin{tabular}{@{}c@{}}
\bfseries Haoyang Jiang$^{1}$, Zhuotong Fan$^{1}$, Mingyang Yi$^{1,\dagger}$, Ju Fan$^{1}$, Lanqing Hu$^{2}$, Xiao Zan$^{3}$\\[5pt]
\normalfont $^{1}$Renmin University of China\\[2pt]
\normalfont $^{2}$Independent Researcher \quad $^{3}$Beike Inc.\\[4pt]
\normalfont\fontsize{9pt}{11pt}\selectfont\texttt{\{jianghaoyang233,fzt86,yimingyang,fanj\}@ruc.edu.cn}\\[2pt]
\normalfont\fontsize{9pt}{11pt}\selectfont\texttt{hulanqing1@outlook.com}\quad\texttt{xiaozan001@ke.com}\\[3pt]
\normalfont $^{\dagger}$Corresponding author
\end{tabular}%
}%
}
\iclrfinalcopy  % Show the author block.

\begin{document}

\maketitle

% Remove the conference-publication claim; retain the original header rule.
\fancyhead{}
\renewcommand{\headrulewidth}{0.4pt}
\pagestyle{fancy}
\thispagestyle{fancy}

\begin{abstract}
Table reasoning with language models typically assumes access to clean textual tables. In practice, tables are often serialized into text, which can corrupt cell contents, distort structural relationships, and discard visual cues, leaving the model to reason over incomplete evidence. In contrast, table images retain this information, but using it requires both accurate perception and reasoning grounded in the relevant cells. To address this, we propose \textsc{Merit} (\textbf{M}ultimodal \textbf{E}vidence \textbf{R}easoning over \textbf{I}mage \textbf{T}ables), a three-stage framework for learning to reason over table images, with serialized tables serving as optional auxiliary inputs (if provided). Our analysis identifies visual evidence acquisition as a bottleneck, motivating table-specific encoder adaptation through reconstruction, localization, and color-conditioned supervision. Building on the adapted representations, supervised fine-tuning teaches explicit evidence selection and reasoning using verified trajectories. Finally, reinforcement learning refines the policy with separate rewards for evidence correctness, calculation, final answers, and output format. By providing feedback on intermediate steps, the framework targets retrieval and computational errors that final-answer supervision alone does not distinguish. We evaluate the approach on table question answering and fact verification, analyze the contributions of visual inputs and training components, and assess generalization to additional datasets with different table structures and visual presentations.
\end{abstract}

% Section 1: motivation and paper overview. Figure 1 is included here.

\section{Introduction}
\label{sec:introduction}

\begin{figure*}[!t]
    \vspace{-14mm}
    \centering
    \includegraphics[width=0.9\textwidth]{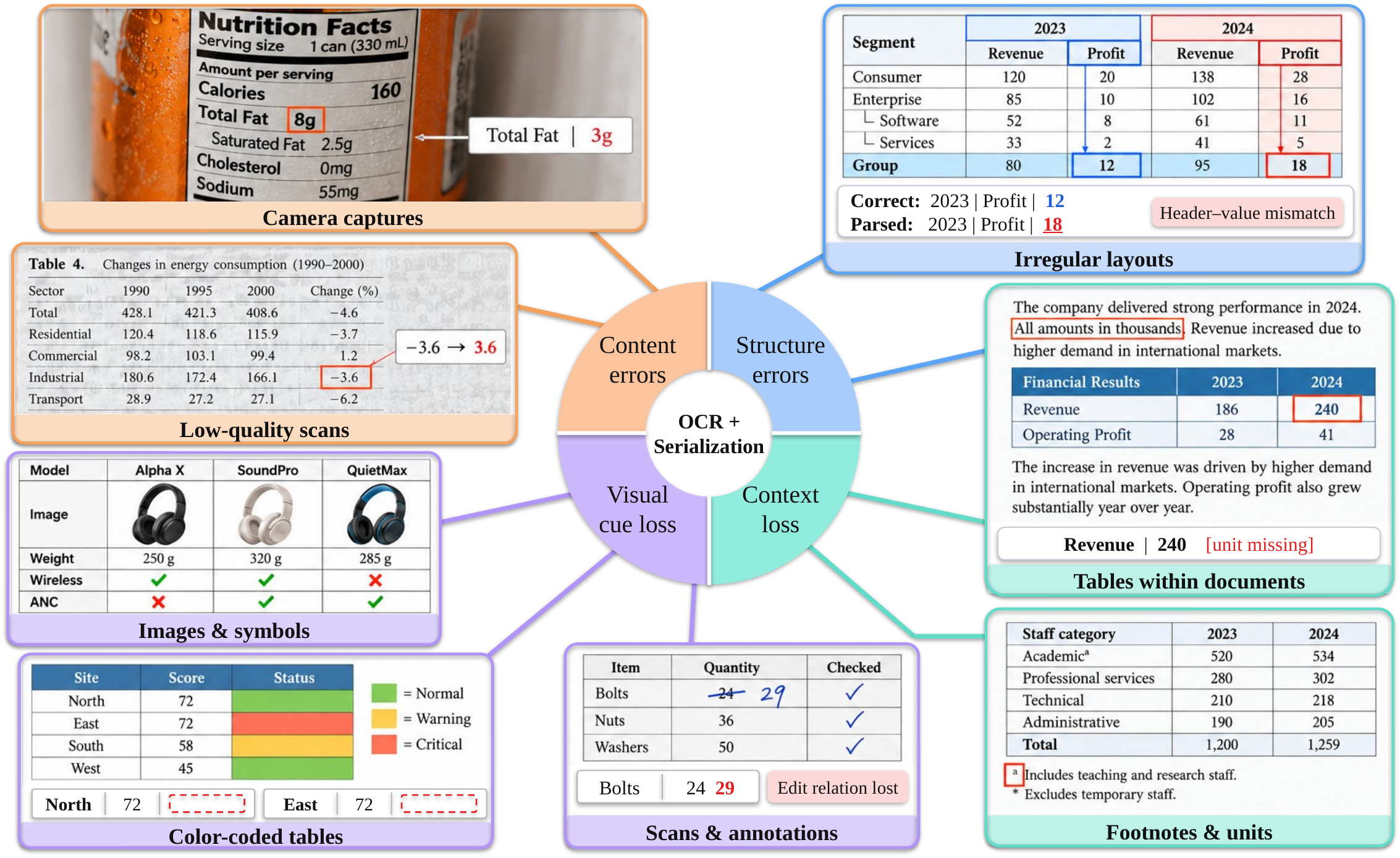}
    \caption{Common failure modes of OCR and table serialization. Converting table images into text can introduce content errors, structure errors, context loss, and visual cue loss, whereas the original table image preserves this evidence.}
    \label{fig:motivation}
    \vspace{-5mm}
\end{figure*}

Recently, with the rapid development of large language models (LLMs) \citep{zhao2026survey,naveed2025overview}, LLMs have become standard tools for table question answering (Table QA), typically taking tables serialized as clean HTML, JSON, or text sequences as input prefixes~\citep{pasupat2015wtq,chen2020tabfact,chen2021finqa,zhang2026reward}. However, tables also appear in PDFs, scans, screenshots, or rendered web pages. These tables are usually converted to text by OCR~\citep{wang2023ocr} and table parsing~\citep{yu2024survey,abdallah2024deep}, and then processed as standard textual tables. In practice, these textual tables are encoded into a one-dimensional sequence. However, as visualized in Figure \ref{fig:motivation}, this conversion can misalign rows and columns, flatten hierarchical headers, omit cells, and discard merged regions or visual cues such as color and shading~\citep{wang2024mineru,li2022ppstructurev2}. A parsed table may remain syntactically valid despite these errors, causing a downstream model to reason confidently over corrupted evidence.

Due to the aforementioned limitations, recognizing tables as images provides a natural way to relax this unrealistic dependency on perfect serialization. The table image preserves content, two-dimensional structure, and visual semantics in a single representation, enabling reasoning when structured text is unavailable and complementing imperfect text when it is available. Motivated by the strong performance of vision-language models (VLMs)~\citep{bai2025qwen3vl}, we propose to use them to tackle Table QA by providing the original table as an image. This image is used as direct/auxiliary information, depending on whether a textual table is accessible. 
% our goal is therefore not to replace accurate text, but to make Table QA reliable when accurate text cannot be assumed.

% Recent work has established the feasibility of understanding tables directly from images~\citep{zheng2024tablellava,zhao2024tabpedia}. Controlled studies further show that neither text nor image is universally superior: their effectiveness varies with table size and question complexity, while combining both indiscriminately is not always optimal~\citep{zhou2025texts}. Other approaches exploit hybrid-modal preference learning or structured tables available during training to improve visual table reasoning~\citep{liu2025hippo,jiang2025turbo}. Together, these studies demonstrate the value of visual evidence while also showing that effectively exploiting it for Table QA remains challenging.

However, feeding table images to a VLM is non-trivial. Most VLMs inherit visual encoders pretrained on generic image--text data, which may struggle with table images that contain dense small text, multi-level headers, merged cells, and visual formatting requiring precise row--column alignment. Therefore, we adapt the visual encoder with table-centric supervision to improve its abilities of color understanding, local reading, and row--column localization. However, encoder adaptation alone does not consistently improve end-to-end Table QA. A model may read individual elements more accurately, yet retrieve the wrong values or fail to complete the required computation. This suggests that improving the visual encoder is necessary but insufficient: the model must also learn how to use table evidence to solve the question.

To address this challenge, we develop a multimodal post-training framework combining supervised fine-tuning (SFT) and reinforcement learning (RL). SFT teaches the model table-specific question answering and structured solution patterns, providing a stable initialization for subsequent optimization. RL then uses table-specific verifiable rewards to improve both intermediate information retrieval and final-answer correctness. This produces a complete training progression from table-aware visual adaptation to reliable multimodal Table QA.

Our contributions are summarized as follows:
\begin{itemize}
\item We adapt the visual encoder into a table-aware encoder with table reconstruction, cell localization, and color recognition objectives.

\item Based on the adapted encoder, we train a multimodal Table QA model with multimodal SFT and table-specific RL to retrieve question-relevant evidence and reason over it.

\item Experiments show that our model improves table perception, evidence retrieval, and final-answer accuracy, even when structured table representations are unavailable or imperfect.

\end{itemize}

% Section 2: structured Table QA, visual/multimodal tables, and post-training.
\section{Related Work}
\label{sec:related_work}

\paragraph{Table Question Answering.}

Table QA research typically assumes accurate table serialization and improves reasoning through semantic parsing, table-aware pretraining, cell selection, aggregation, and executable programs~\citep{pasupat2015wtq,yin2020tabert,herzig2020tapas,liu2022tapex,zhang2026reward}. Subsequent studies extend this setting to fact verification and numerical reasoning while improving transparency through explicit row and cell selection~\citep{chen2020tabfact,chen2021finqa,gupta2022right,wu2023tacr}. However, these methods cannot directly recover information lost or corrupted during serialization, whereas our method grounds evidence and reasoning in the original table image. Recent studies also use table images with VLMs for Table QA~\citep{zheng2024tablellava,zhao2024tabpedia,park2025fintabllava,jiang2025cocotab,kim2026tabflash,zhang2025tablemoe}. Complementary studies examine how VLMs use table images and textual representations together for question answering~\citep{zhou2025texts,zhou2025ritt,wang2026hippo,jiang2025turbo}. Document-oriented models improve visual text perception through OCR-free pretraining or document-parsing supervision~\citep{kim2022donut,lee2023pix2struct,niu2026mineru25}, but target general text extraction rather than table structure and visual attributes such as color. In contrast, our framework separates table-aware visual encoder adaptation from subsequent QA post-training, enabling the model to use visual evidence without assuming reliable structured input.

\paragraph{Post-Training for Multimodal Table Reasoning.}

Post-training improves instruction following and reasoning through demonstrations and feedback~\citep{wang2025tandem,su2025reveal,li2026ets,li2026towards,li2026reasoning,zhang2026htam,li2026turning}, and has been applied to both structured and visual table reasoning~\citep{yang2025tabler1,wang2026hippo,jiang2025turbo}. Table-oriented approaches strengthen evidence acquisition and reasoning through task-specific training~\citep{wu2026regiontabler1,liu2026mmtabler1,kang2025tabler1,liu2026tablemix}, with intermediate verification supplementing final-answer supervision~\citep{nguyen2026coretab,jiang2026vtabler1}. Unlike these methods, our framework specifically distinguishes retrieving the correct table evidence from correctly computing it. Concretely, our SFT establishes explicit cell retrieval and calculation, while RL applies deterministic, decomposed rewards to evidence correctness, calculation execution, final-answer correctness, and output format. These designs provide targeted feedback on retrieval and computational errors.

\section{Preliminaries}
\label{sec:preliminaries}

In this section, we briefly introduce VLMs, SFT, and GRPO, which form the foundations of our multimodal post-training framework.

\paragraph{Vision-Language Modeling.}
A VLM typically consists of two critical components: a visual encoder $f_{\phi}$ and an autoregressive language model $\pi_{\theta}$. Given an image $I$ and a textual context $\mathbf{x} = (x_{1}, \ldots, x_{K})$, the model generates a response $\mathbf{y}=(y_1,\ldots,y_T)$
\begin{equation}
\pi_{\theta}(\mathbf{y}\mid I, \mathbf{x})
=
\prod_{t=1}^{T}
\pi_{\theta}
\left(
y_t
\mid
\mathbf{y}_{<t}, f_{\phi}(I),
\mathbf{x}
\right),
\end{equation}
where $\phi$ and $\theta$ denote the parameters of the visual encoder and the language decoder, respectively. In our setting, $I$ is a table image and $\mathbf{x} = [\mathbf{q};\mathbf{s}]$ contains a question $\mathbf{q}$ and an optional textual table $\mathbf{s}$, with $\mathbf{x}=\mathbf{q}$ when structured input is unavailable.

\paragraph{Supervised Fine-Tuning.}
SFT trains pretrained models to follow task instructions, produce responses in a prescribed format, and improve their task-specific capabilities~\citep{ouyang2022training,liu2023visual}. Given a supervised dataset
$\mathcal{D}_{\mathrm{SFT}}
=\{(I_i,\mathbf{x}_i,\mathbf{y}_i^*)\}_{i=1}^{N}$,
where $\mathbf{y}_i^*=(y_{i,1}^*,\ldots,y_{i,T_i}^*)$ is the reference response containing $T_i$ tokens, SFT minimizes the token-level negative log-likelihood:
\begin{equation}
\mathcal{L}_{\mathrm{SFT}}(\phi,\theta)
=
-\frac{1}{\sum_{i=1}^{N}T_i}
\sum_{i=1}^{N}
\sum_{t=1}^{T_i}
\log
\pi_{\theta}
\left(
y_{i,t}^*
\mid
\mathbf{y}_{i,<t}^*,
f_{\phi}(I_i),
\mathbf{x}_i
\right).
\end{equation}
In our framework, SFT teaches the model to retrieve table evidence and present its calculations and answer in a parseable format. This lets RL refine these components separately, rather than relying on reward feedback alone to teach the solution procedure and output format.

\paragraph{Group Relative Policy Optimization.}
GRPO is a policy-optimization algorithm that avoids training a separate value model by estimating advantages from the relative rewards of multiple responses sampled for the same input~\citep{shao2024deepseekmath}. Concretely, given an input $\rvx$ and image $I$, GRPO samples a group of $G$ responses $\{\mathbf{y}_i\}_{i=1}^{G}$ from the old policy $\pi_{\theta_{\mathrm{old}}}$ and assigns each response a scalar reward $s_i=\mathcal{R}(\rvx,\mathbf{y}_i)$. The group-relative advantage is computed as
\begin{equation}
    \widehat{A}_i
    =
    \frac{
        s_i-\operatorname{mean}\!\left(\{s_j\}_{j=1}^{G}\right)
    }{
        \operatorname{std}\!\left(\{s_j\}_{j=1}^{G}\right)+\delta
    },
\end{equation}
where $\delta$ is a small constant for numerical stability. GRPO then maximizes a clipped surrogate objective with a KL penalty relative to a reference policy, taking the expectation over inputs and response groups sampled from the old policy:
\begingroup
\small
\begin{equation}\label{eq:grpo_objective}
\begin{aligned} \mathcal{J}_{\mathrm{GRPO}}(\theta) &= \mathbb{E}\Bigg[ \frac{1}{G}\sum_{i=1}^{G} \frac{1}{|\mathbf{y}_i|} \sum_{t=1}^{|\mathbf{y}_i|} \Big\{ \min\big( \rho_{i,t}(\theta)\widehat{A}_i,\, \operatorname{clip}( \rho_{i,t}(\theta),1-\epsilon,1+\epsilon )\widehat{A}_i \big) \\ &\qquad -\beta D_{\mathrm{KL}}\big( \pi_{\theta}(\cdot\mid h_{i,t}) \,\Vert\, \pi_{\mathrm{ref}}(\cdot\mid h_{i,t}) \big) \Big\} \Bigg], \end{aligned}
\end{equation}
\endgroup
where $h_{i,t}=(f_{\phi}(I),\rvx,\mathbf{y}_{i,<t})$ and
$\rho_{i,t}(\theta)=
\pi_{\theta}(y_{i,t}\mid h_{i,t})/
\pi_{\theta_{\mathrm{old}}}(y_{i,t}\mid h_{i,t})$.
Here, $\epsilon$ and $\beta$ control clipping and KL regularization, respectively. Section~\ref{sec:table_rl} specifies the table-specific reward $\mathcal{R}$.

% Section 3: encoder adaptation, Structured Selection SFT, and verifiable RL.
% Figure 2 remains a compilation-safe placeholder until the framework diagram
% is exported as figures/fig2_framework.pdf.

\begin{figure*}[t]
    \vspace{-14mm}
    \centering
    \includegraphics[width=0.90\textwidth]{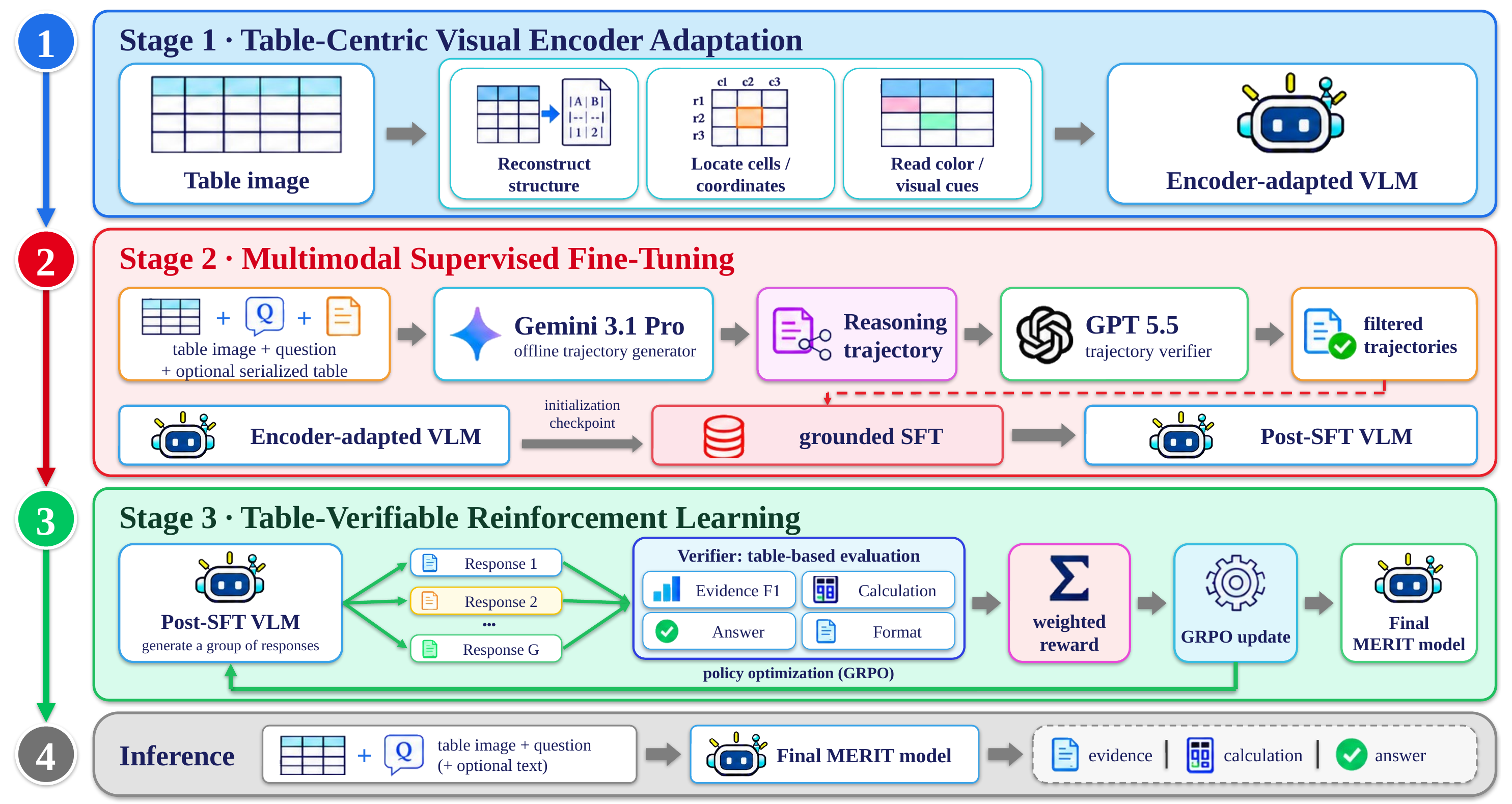}
    \caption{Overview of \textsc{Merit}. Table-centric visual encoder adaptation first specializes the visual encoder for table images. Multimodal SFT then teaches explicit evidence selection and structured reasoning, and table-verifiable RL further improves evidence and answer accuracy with verifiable rewards.}
    \label{fig:framework}
    \vspace{-8pt}
\end{figure*}

\section{Method}
\label{sec:method}

As discussed in Section~\ref{sec:introduction}, real-world tables may be image-only or lose content, structure, and visual semantics during OCR and parsing. \textsc{Merit} therefore uses table images as primary input and text as optional auxiliary information. Figure~\ref{fig:framework} shows its three stages linking perception and verifiable reasoning: \textbf{Table-Centric Visual Encoder Adaptation} specializes the vision backbone for dense cell content, grid coordinates, and visual attributes; \textbf{Multimodal SFT} teaches question-relevant evidence retrieval and grounded reasoning; and \textbf{Table-Verifiable RL} optimizes reasoning with targeted rewards for evidence retrieval, calculation, and answer correctness.

\subsection{Table-Centric Visual Encoder Adaptation}
\label{sec:visual_adaptation}

Table images provide primary evidence when structured tables are unavailable and complementary evidence when serialized tables are incomplete or noisy. Their utility depends on the visual encoder's ability to extract image information. However, VLMs such as Qwen3-VL~\citep{bai2025qwen3vl} are mainly pretrained on broad image--text data, while tables contain dense text, regular grids, hierarchical headers, merged cells, and visual attributes requiring precise row--column alignment. We therefore diagnose this perceptual bottleneck before introducing targeted table supervision for the visual encoder.

\begin{figure*}[t]
    \vspace{-14mm}
    \centering
    % Left panel: diagnostic table
    \begin{subfigure}[c]{0.36\textwidth}
        \centering
        \renewcommand{\arraystretch}{1.15}
        \small
        \resizebox{\linewidth}{!}{%
        \begin{tabular}{lc}
            \toprule
            \textbf{Setting / Visual Encoder} & \textbf{Acc. (\%)} \\
            \midrule
            \multicolumn{2}{l}{\textit{NaViT \(\rightarrow\) Qwen3 decoder}} \\
            \quad Normal & 57.8 \\
            \quad +Gold Facts & 79.8 \\
            \midrule
            \multicolumn{2}{l}{\textit{MinerU 2.5 \(\rightarrow\) Qwen3 decoder}} \\
            \quad Normal & 72.2 \\
            \quad +Gold Facts & 79.6 \\
            \midrule
            \textbf{Gain from doc/table training} & \\
            \quad Normal & \(+14.4\) \\
            \quad +Gold Facts & \(-0.2\) \\
            \bottomrule
        \end{tabular}%
        }

        \vspace{2pt}

        \caption{Accuracy (\%) under different visual encoders.}
        \label{tab:encoder_diagnostic_table}
    \end{subfigure}
    \hfill
    % Right panel: representative image-matters case
    \begin{subfigure}[c]{0.60\textwidth}
        \centering
        \IfFileExists{iclr2027/figs/image_matters_case.pdf}{%
            \includegraphics[
                width=\linewidth,
                height=5.2cm,
                keepaspectratio
            ]{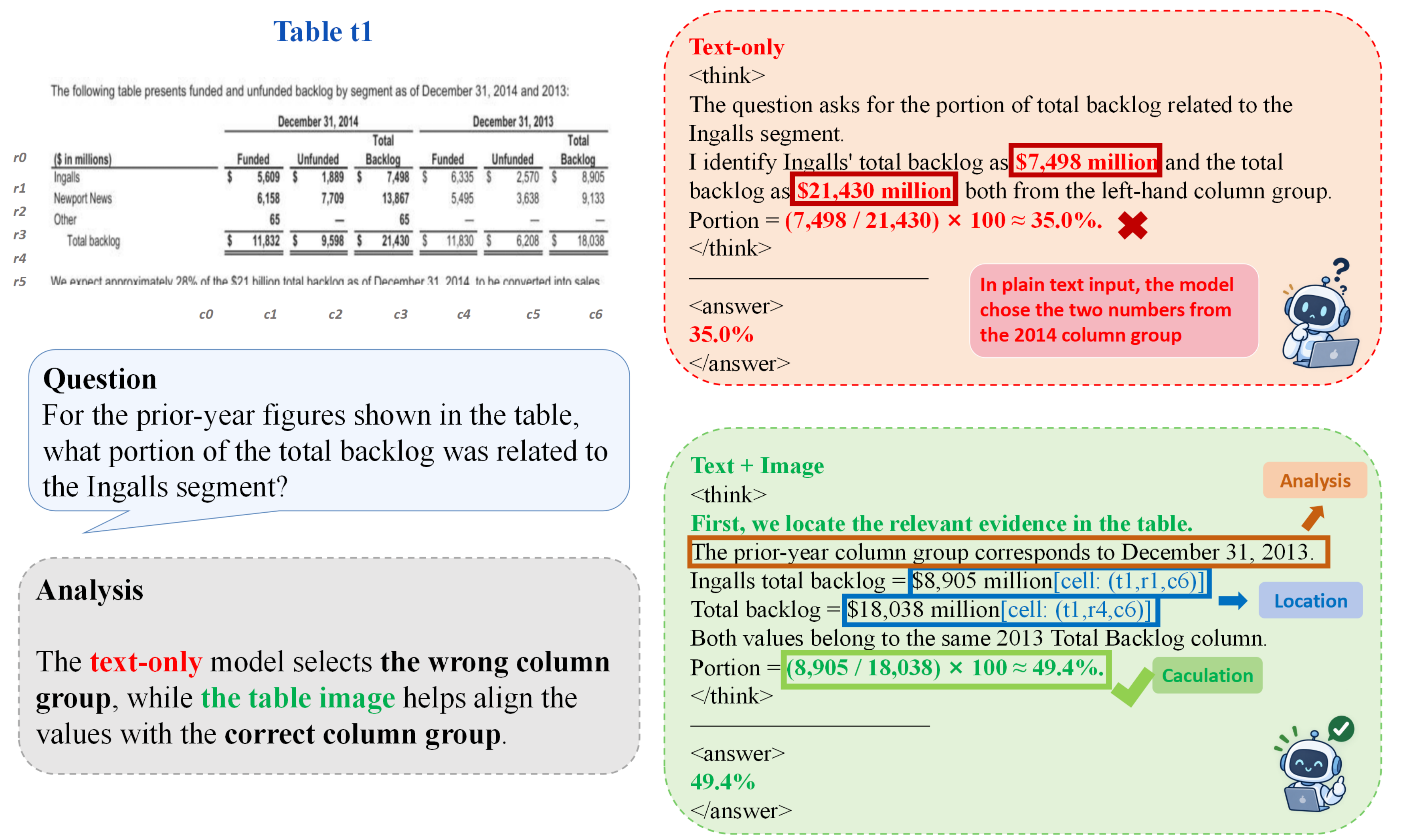}%
        }{%
            \fbox{%
                \parbox[c][5.0cm][c]{0.94\linewidth}{%
                    \centering
                    \textbf{Image Matters Case}\\[4pt]
                    Place the example at\\
                    \texttt{iclr2027/figs/image\_matters\_case.pdf}
                }%
            }%
        }
        \caption{Text-only input selects values from the wrong column group, whereas the table image aligns them with the correct one.}
        \label{fig:image_matters_case}
    \end{subfigure}

    \caption{
        Diagnosing the visual bottleneck in table question answering.
        (a) NaViT and MinerU~2.5 are connected to the same Qwen3 decoder.
        The accuracy gap under normal conditions largely disappears when
        gold facts are provided, indicating that the visual encoder is the
        primary bottleneck. (b) An example in which the table image corrects
        an evidence-selection error made from text-only input.
    }
    \label{fig:encoder_diagnostic}
    \vspace{-7mm}
\end{figure*}

\paragraph{Visual-encoder diagnostic.}
We conduct a controlled experiment on the Table QA benchmark FinQA~\citep{chen2021finqa}, coupling either the NaViT encoder of Qwen2-VL or MinerU~2.5~\citep{niu2026mineru25} with the same decoder. MinerU~2.5 builds on this encoder with additional document- and table-oriented training for visual text perception. As shown in Figure~\ref{tab:encoder_diagnostic_table}, it raises image-based accuracy from $57.8\%$ to $72.2\%$, a $14.4$-point gain. With gold supporting facts, the models perform nearly identically ($79.8\%$ versus $79.6\%$). Because the decoder is fixed and gold facts bypass visual evidence extraction, the gain appears to stem primarily from better evidence perception rather than reasoning over provided evidence. This finding motivates our table-specific visual encoder adaptation.\footnote{We do not directly adopt MinerU~2.5: connecting its encoder to the language decoder would still incur comparable alignment costs, and its training emphasizes text extraction rather than visual attributes such as color.}

\paragraph{Targeted adaptation.}
Our analysis of failure cases (Appendix~\ref{app:encoder_badcase}) shows that, when processing table images, visual encoders commonly misread dense cell content, associate values with incorrect row--column positions, or miss question-relevant visual attributes. These three perceptual errors motivate the following task families:
\begin{equation}
    \mathcal{K}
    =
    \{
        \textsc{Recon},
        \textsc{Locate},
        \textsc{Color}
    \}.
    \label{eq:adaptation_tasks}
\end{equation}
\textsc{Recon} reconstructs textual tables from rendered images, supporting dense-content recognition and structural preservation. \textsc{Locate} uses cell-, row-, and column-level queries to map visual content to logical grid positions. \textsc{Color} trains visual-attribute grounding by assigning random background colors to selected cells and asking questions whose answers depend on those assignments. Since colors are sampled independently of cell values, answering requires visual rather than textual cues.

All three tasks use response-generation objectives. The training data are constructed from MMTab~\citep{zheng2024tablellava}, together with auxiliary table QA examples from MMTab. Reconstruction and localization targets come from source annotations, whereas color-conditioned targets are generated from assigned colors and cell contents. We compute a length-normalized autoregressive cross-entropy for each task family and combine the losses using task-specific weights:
\begin{equation}
    \min_{\phi}\mathcal{L}_{\mathrm{adapt}}(\phi)
    =
    \sum_{k\in\mathcal{K}}
    \lambda_k \mathcal{L}_k(\phi),
    \qquad
    \mathcal{L}_k(\phi)
    =
    -\frac{1}{|\mathbf{y}|}
    \sum_{t=1}^{|\mathbf{y}|}
    \log
    \pi_{\theta}
    \left(
        y_t
        \mid f_{\phi}(I),X_k,\mathbf{y}_{<t}
    \right),
    \label{eq:visual_adaptation}
\end{equation}
where $I$ denotes the table image, $X_k$ denotes the task-specific textual input for task family $k$, and $\mathbf{y}$ denotes the target response. The visual encoder is trained by minimizing the weighted loss in Eq.~\eqref{eq:visual_adaptation} over the collected data. Data construction and instruction templates are detailed in Appendix~\ref{app:visual_adaptation_data}.

\paragraph{Effect of encoder adaptation.}
Table~\ref{tab:heldin} evaluates the adapted encoder on six table-structure understanding tasks. The proposed adaptation improves the macro-average from $46.89$ to $53.51$, with particularly strong gains in cell extraction, cell localization, and row--column extraction. These results confirm that the targeted objectives strengthen the table-specific perceptual capabilities required by downstream reasoning.

\begin{table*}[b]
    \vspace{-5mm}
\centering
\caption{
Evaluation results on six held-in table-structure understanding benchmarks from MMTab~\citep{zheng2024tablellava}.
Metric definitions follow~\citet{zheng2024tablellava} (see Section~\ref{app:structure_eval}).
For benchmarks with multiple submetrics, we first average their submetrics; Avg.\ then denotes the macro-average over the six benchmarks.
Higher values indicate better performance.
}
\label{tab:heldin}
\resizebox{\textwidth}{!}{%
\begin{tabular}{llccccccccccc}
\toprule
\multirow{3}{*}{\textbf{Method}}
& \multirow{3}{*}{\textbf{LLM}}
& \multicolumn{2}{c}{\textbf{TSD}}
& \textbf{TCE}
& \textbf{TCL}
& \textbf{MCD}
& \multicolumn{2}{c}{\textbf{RCE}}
& \multicolumn{3}{c}{\textbf{TR}}
& \multirow{3}{*}{\textbf{Avg.}} \\
\cmidrule(lr){3-4}
\cmidrule(lr){5-5}
\cmidrule(lr){6-6}
\cmidrule(lr){7-7}
\cmidrule(lr){8-9}
\cmidrule(lr){10-12}
& & Row & Col.
& \multirow{2}{*}{Acc.}
& \multirow{2}{*}{Acc.}
& \multirow{2}{*}{F1}
& Row & Col.
& HTML & Markdown & LaTeX
& \\
& & Acc. & Acc.
& & &
& F1 & F1
& TEDS & TEDS & TEDS
& \\
\midrule

% Table-LLaVA 7B
% & Vicuna-1.5 7B
% & 33.10 & 33.20
% & 19.45 & 29.31 & 17.14
% & 31.43 & 37.93
% & 50.24 & 44.82 & 46.11
% & 30.13 \\

Table-LLaVA 7B
& Vicuna-1.5 7B
& 41.50 & 59.00
& 27.68 & 41.67 & 38.85
& 38.45 & 48.67
& 63.88 & 66.58 & 66.52
& 44.61 \\

% Table-LLaVA 13B
% & Vicuna-1.5 13B
% & 34.40 & 27.60
% & 19.53 & 29.68 & 16.52
% & 31.07 & 41.49
% & 51.44 & 46.00 & 46.50
% & 30.17 \\

Table-LLaVA 13B
& Vicuna-1.5 13B
& 44.40 & 75.00
& 30.88 & 45.49 & \textbf{43.60}
& 37.74 & 55.02
& \textbf{66.24} & 66.75 & \textbf{66.69}
& 48.77 \\

Base
& Qwen3-VL 8B-Instruct
& 37.80 & 72.40
& 37.46 & 38.63 & 30.99
& 41.19 & 68.02
& 59.26 & 88.19 & 46.25
& 46.89 \\

\textbf{Ours}
& Qwen3-VL 8B-Instruct
& \textbf{45.20} & \textbf{73.20}
& \textbf{45.56} & \textbf{48.75} & 35.91
& \textbf{51.40} & \textbf{75.35}
& 62.12 & \textbf{89.96} & 52.76
& \textbf{53.51} \\

\bottomrule
\end{tabular}
}
\vspace{-7mm}
\end{table*}

Notably, though these auxiliary tasks improve table perception, they do not directly train the complete evidence-to-answer procedure required for downstream QA. Therefore, we next use multimodal SFT to initialize table reasoning and establish the output protocol for subsequent reinforcement learning.

\subsection{Multimodal Supervised Fine-Tuning}
\label{sec:multimodal_sft}

The adapted encoder provides table-specific visual representations but does not teach the model to select question-relevant evidence, perform calculations, or produce parseable responses. Applying RL directly would require learning both evidence-grounded reasoning and the output protocol from sparse rewards. We therefore initialize the adapted model with multimodal SFT on grounded demonstrations. Each demonstration pairs a table image, optional textual context, and a question with a verified trajectory specifying supporting evidence, calculations when needed, and the final answer. SFT establishes initial table-solving behavior and a standard output format for component-wise verifiable rewards in subsequent RL. Data construction and composition are described below and in Appendix~\ref{app:sft_data_composition}.

\paragraph{Grounded trajectory construction.}
We use Gemini-3.1-Pro~\citep{google2026gemini31pro} to generate answer-conditioned Chain-of-Thought (CoT) trajectories~\citep{wei2022cot} from table images, structured tables, cell metadata, questions, and annotated answers. Each trajectory identifies relevant headers and cells, recording each cited cell as $(r,c,v)$, where $r$ and $c$ are its logical row and column and $v$ its normalized value. It checks cited values against the corresponding cells before lookup, comparison, calculation, or factual inference, retaining visual conditions such as color when they affect evidence selection. GPT-5.5~\citep{openai2026gpt55} then independently verifies the cited tuples against the source inputs, as well as the derivation and the final answer. Only trajectories passing all three checks enter $\mathcal{D}_{\mathrm{SFT}}$. Each accepted trajectory therefore provides
\begin{equation}
    \left(
        E^{\mathrm{ref}},
        z^{\mathrm{ref}},
        A^{*}
    \right),
    \qquad
    E^{\mathrm{ref}}
    =
    \left\{
        (r_j,c_j,v_j)
    \right\}_{j=1}^{M},
    \label{eq:sft_structured_targets}
\end{equation}
where $E^{\mathrm{ref}}$ contains the verified evidence tuples, $z^{\mathrm{ref}}$ the verified calculation when applicable, and $A^{*}$ the annotated answer or verification label.

\begin{figure}[t]
\vspace{-7mm}
    \centering
    \includegraphics[
        width=0.86\linewidth,
        trim={8pt 20pt 8pt 6pt},
        clip
    ]{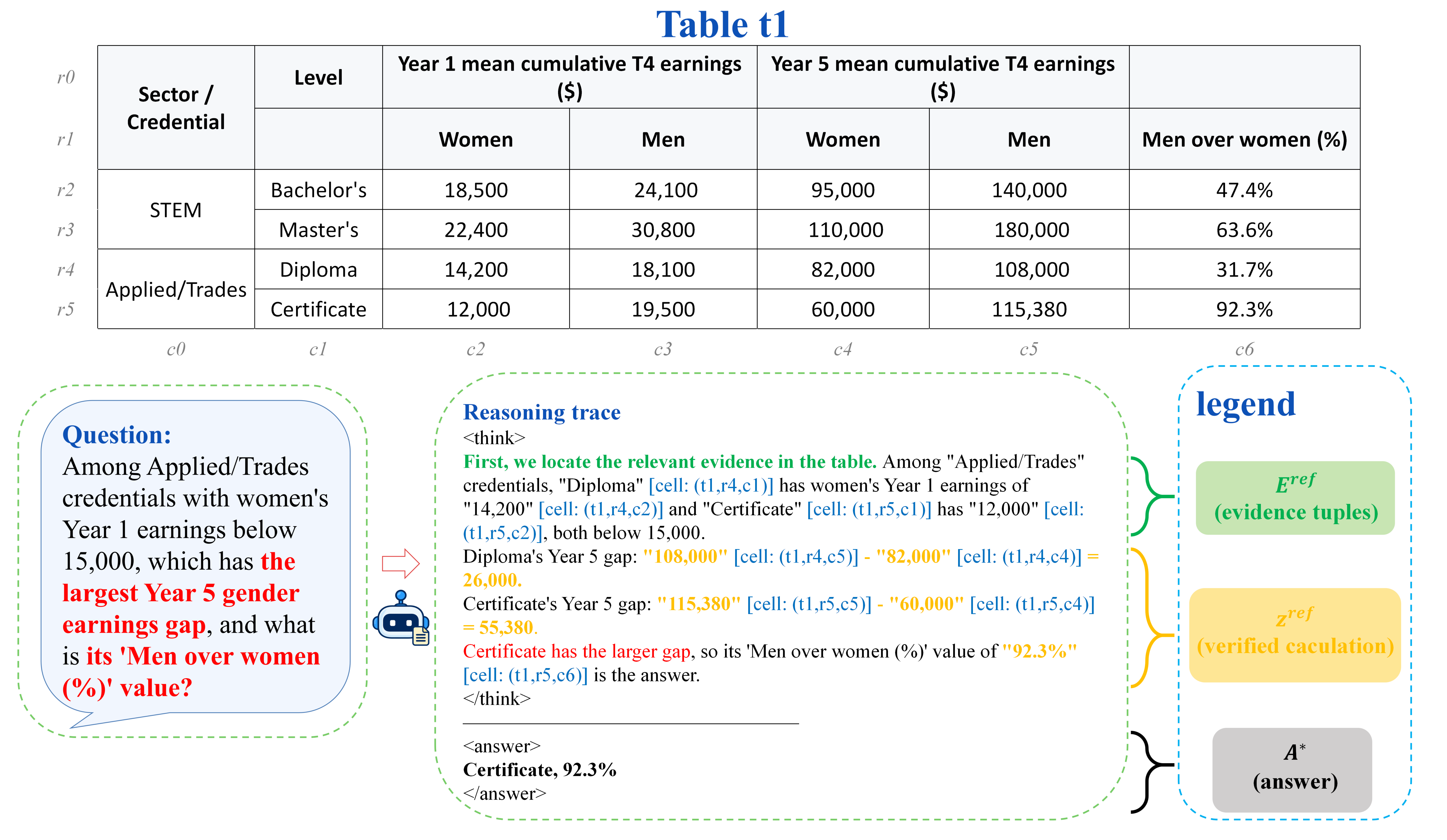}
    \vspace{-1.5mm}
    \caption{
    Illustration of a grounded reasoning trajectory. The model identifies supporting cells using logical coordinates, performs the required calculation, and produces the final answer in a standardized format. The example is illustrative, and intermediate steps are abbreviated for clarity.
    }
    \label{fig:cot_case}
    \vspace{-7mm}
\end{figure}

The response serializes the trajectory $R(E^{\mathrm{ref}},z^{\mathrm{ref}})$ in a unified format:
\begin{equation}
    \mathbf{y}^{*}
    =
    \thinkopen\,
    R(E^{\mathrm{ref}},z^{\mathrm{ref}})\,
    \thinkclose\,
    \answeropen\,
    A^{*}\,
    \answerclose.
    \label{eq:sft_response_format}
\end{equation}
This format exposes evidence, calculation, and answer as separately parseable components. Figure~\ref{fig:cot_case} illustrates how the selected table cells are linked to the intermediate calculation and final answer.

\paragraph{Training objective.}

For each accepted response $\mathbf{y}^{*}$ and the corresponding table image $I$ and query $\rvx$, we minimize the length-normalized autoregressive cross-entropy of the policy $\pi_{\theta}$ (the SFT objective in Section~\ref{sec:preliminaries}) with the visual encoder $f_{\phi}$ frozen.  
% \begin{equation}
%     \min_{\theta}\mathcal{L}_{\mathrm{SFT}}
%     =
%     -\frac{1}{|\mathbf{y}^{*}|}
%     \sum_{t=1}^{|\mathbf{y}^{*}|}
%     \log
%     \pi_{\theta}
%     \left(
%         y_t^{*}
%         \mid f_{\phi}(I),X,\mathbf{y}_{<t}^{*}
%     \right),
%     \label{eq:multimodal_sft}
% \end{equation}
%where $I$ denotes the table image, and $X$ contains the question and any accompanying textual context. Here, $\phi$ and $\theta$ denote the vision encoder and language decoder parameters, respectively. The bar indicates frozen parameters. Thus, SFT keeps the adapted vision encoder $\bar{\phi}$ fixed and updates only the language decoder $\theta$. 
The training loss is computed only over response tokens, thereby supervising both the grounded reasoning trajectory and the standardized output format. Generation and verification prompts are provided in Appendix~\ref{app:generation_verification_prompts}.

SFT establishes an initial table-solving policy and a parseable generation protocol, but it only imitates fixed demonstrations under teacher forcing. We therefore further optimize responses sampled from the policy using table-verifiable reinforcement learning.

\subsection{Table-Verifiable Reinforcement Learning}
\label{sec:table_rl}

After SFT, the policy follows a standardized output format, but its likelihood objective does not evaluate the correctness of sampled responses. We therefore further improve the SFT model with GRPO using separate rewards for evidence retrieval $r_{\mathrm{evi}}$, calculation $r_{\mathrm{calc}}$, final-answer correctness $r_{\mathrm{ans}}$, and output format $r_{\mathrm{fmt}}$. Following Section~\ref{sec:multimodal_sft}, we extract the cited cells, arithmetic expression, and final answer from each generated response.

The total reward is a weighted sum:
\begin{equation}
    r
    =
    \lambda_{\mathrm{evi}}r_{\mathrm{evi}}
    +
    \lambda_{\mathrm{calc}}r_{\mathrm{calc}}
    +
    \lambda_{\mathrm{ans}}r_{\mathrm{ans}}
    +
    \lambda_{\mathrm{fmt}}r_{\mathrm{fmt}},
    \label{eq:total_reward}
\end{equation}
where the nonnegative coefficients $\lambda$ control the contribution of each component. We use $r$ as the reward $\mathcal{R}$ in the GRPO objective~\eqref{eq:grpo_objective}.

\paragraph{Verifiable reward components.}
The evidence reward measures the set-level F1 between the predicted and
reference evidence cells:
\begin{equation}
    r_{\mathrm{evi}}
    =
    \frac{2N_{\mathrm{match}}}
    {N_{\mathrm{pred}}+N_{\mathrm{ref}}},
    \label{eq:evidence_reward}
\end{equation}
where $N_{\mathrm{pred}}$ and $N_{\mathrm{ref}}$ are the numbers of
predicted and reference evidence cells, respectively, and
$N_{\mathrm{match}}$ is the number of correctly matched cells. A
predicted cell is counted as matched only when its logical row, column,
and normalized value all agree with the same reference cell.

For arithmetic questions, let $\hat{z}$ denote the calculation extracted
from the model response and $z^{\mathrm{ref}}$ the verified reference
calculation. The calculation reward is the binary indicator
$\mathbb{I}_{\{\hat{z}=z^{\mathrm{ref}}\}}$, where equality is determined
by executing both calculations and comparing their normalized numerical
results. Numerical normalization removes formatting symbols and converts
percentages and units when necessary. For non-arithmetic questions,
$\lambda_{\mathrm{calc}}$ is set to zero.

Let $\hat{A}$ denote the answer extracted from the model response and
$A^{*}$ the annotated answer. The answer reward is the binary indicator
$\mathbb{I}_{\{\hat{A}=A^{*}\}}$, where the comparison is performed after
normalizing capitalization, whitespace, punctuation, and numerical
formatting.

Finally, the format reward is the binary indicator
$\mathbb{I}_{\{\mathrm{valid\ format}\}}$. It equals one when all required
fields can be parsed from the model response and zero otherwise. The
required fields include the evidence and final answer for all questions,
as well as the calculation for arithmetic questions.

For each input, we score a group of sampled responses using \eqref{eq:total_reward} and update the policy with the GRPO objective in Section~\ref{sec:preliminaries}. The SFT checkpoint initializes both the trainable policy and the fixed reference policy. We keep the adapted visual encoder frozen and update the language decoder. Reward weights, parsing rules, and numerical tolerances are specified in Appendix~\ref{app:reward_parsing_details}.

% Section 4: current measured results plus explicit TBD slots for unfinished runs.
%\input{experiments}

\providecommand{\cmark}{\ensuremath{\checkmark}}
\providecommand{\xmark}{\ensuremath{\times}}
\providecommand{\na}{\textnormal{N/A}}

\section{Experiments}
\label{sec:experiments}

In this section, we evaluate our framework against existing methods on table question answering and fact verification to determine whether the information preserved in table images improves table reasoning. Furthermore, we evaluate its generalization to out-of-distribution (OOD) data, particularly when serialized tables are unavailable. Finally, ablation studies examine the contribution of each training stage.

% We examine how this value depends on the proposed training procedure through ablations of visual adaptation, SFT, and RL, together with a controlled comparison of input representations. The resulting ability to read table images and reason over their content should also support transfer to new tables. We test this generalization on additional datasets with different table layouts and visual presentations, including settings without serialized tables.

\paragraph{Implementation.}
We initialize from Qwen3-VL-8B-Instruct~\citep{bai2025qwen3vl} and apply visual-encoder adaptation, multimodal SFT, and GRPO as described in Section~\ref{sec:method}. Inputs comprise table images, questions or statements, and accompanying passages; serialized tables are supplied only in the specified input comparisons. Data construction and training configurations are detailed in Appendix~\ref{app:training_data_details}.

\paragraph{Evaluation.}
\label{sec:experimental_setup}
We consider two tasks. Table question answering includes TABMWP~\citep{lu2023tabmwp} for mathematical word problems, WikiTableQuestions (WTQ)~\citep{pasupat2015wtq} for compositional reasoning, and TAT-QA~\citep{zhu2021tatqa} for financial table-and-text reasoning. Table fact verification includes TabFact~\citep{chen2020tabfact} with binary labels and InfoTabs~\citep{gupta2020infotabs} with three-way labels. Following HIPPO~\citep{wang2026hippo} and TURBO~\citep{jiang2025turbo}, we train SFT and GRPO on these five datasets and evaluate on the same held-out splits with the same metrics. To assess generalization, we further evaluate on HiTab~\citep{cheng2022hitab}, FinQA~\citep{chen2021finqa}, FinMMR~\citep{tang2025finmmr}, TableVista~\citep{yang2026tablevista}, and MMMU-T~\citep{yue2024mmmu} without further fine-tuning; these datasets vary in table structure and visual presentation, and only HiTab and FinQA provide serialized tables; for the other three, we convert table images into Markdown with MinerU~2.5~\citep{niu2026mineru25}, and visual models receive both the image and this text. We report accuracy and unweighted averages. Dataset, split, and scoring details are in Appendix~\ref{app:evaluation_details}.

\begin{table*}[t!]
    \vspace{-13mm}
\centering
\caption{
Effectiveness on held-out splits of the five datasets used for reasoning
post-training. Scores are accuracy (\%); Avg.\ is the unweighted mean
across the five datasets. The three rows for our model report results
with different table inputs, indicated in parentheses. \(\dagger\)
denotes results taken from published reports. Answers are extracted and
evaluated against the ground truth, with ambiguous cases reviewed by
GPT-5.5.
}
\label{tab:main_results}
\setlength{\tabcolsep}{6pt}
\renewcommand{\arraystretch}{0.95}
\small
\begin{tabular}{lcccccc}
\toprule
\multirow{2}{*}{\textbf{Method}}
& \multicolumn{3}{c}{\textbf{Table Question Answering}}
& \multicolumn{2}{c}{\textbf{Fact Verification}}
& \multirow{2}{*}{\textbf{Avg.}} \\
\cmidrule(lr){2-4}
\cmidrule(lr){5-6}
& TABMWP & WTQ & TAT-QA
& TabFact & InfoTabs
& \\
\midrule

\multicolumn{7}{l}{\textit{Proprietary models}} \\
GPT-5.5
& 89.97 & \textbf{88.96} & 85.62 & \textbf{93.65} & 76.59 & \textbf{86.96} \\
Claude Sonnet 5
& 89.97 & 82.94 & 88.29 & 81.94 & 74.58 & 83.55 \\
Gemini 3.5 Flash
& 84.62 & \underline{87.63} & \underline{88.63} & \underline{92.64} & 78.93 & 86.49 \\

\midrule

\multicolumn{7}{l}{\textit{General-purpose models}} \\
InternVL2.5-8B
& 93.56 & 54.77 & 55.18 & 72.49 & 67.30 & 68.66 \\
Qwen3-VL-8B
& 85.97 & 61.04 & 82.51 & 77.71 & 73.04 & 76.05 \\
Qwen3-8B
& 86.20 & 76.24 & 80.18 & 86.22 & 79.19 & 81.61 \\

\midrule
\multicolumn{7}{l}{\textit{Table-specialized models}} \\
Table-LLaVA-7B\(^\dagger\)
& 53.20 & 16.62 & 10.49 & 57.62 & 66.78 & 40.94 \\
TabPedia\(^\dagger\)
& 10.66 & 23.53 & 13.08 & 35.49 & 2.43 & 17.04 \\
HIPPO\(^\dagger\)
& 87.34 & 55.71 & 61.40 & 82.29 & 75.70 & 72.49 \\
TURBO\(^\dagger\)
& 96.75 & 67.80 & 73.21 & 85.81 & \textbf{81.89} & 81.09 \\

\midrule
\multicolumn{7}{l}{\textit{Text-only RL baseline}} \\
Qwen3-8B + RL
& 92.85 & 73.31 & 84.95 & 85.58 & 76.39 & 82.62 \\

\midrule
\multicolumn{7}{l}{\textit{Our model}} \\
Ours (Text only)
& \underline{98.48} & 64.71 & \textbf{90.16}
& 87.95 & 75.80 & 83.42 \\

Ours (Image only)
& 98.13 & 72.86 & 87.56
& 90.08 & 78.63 & 85.45 \\

\textbf{Ours (Image + Text)}
& \textbf{98.69} & 76.40 & 87.44
& 90.84 & \underline{80.15} & \underline{86.70} \\

\bottomrule
\end{tabular}
\vspace{-5mm}
\end{table*}

% \paragraph{Experimental Setup.}
% We evaluate our method on table question answering and table fact
% verification. The table question answering datasets include TABMWP
% , WikiTableQuestions (WTQ)
% , and TAT-QA. The table fact
% verification datasets include TabFact and
% InfoTabs. Our model is initialized from
% Qwen3-VL-8B-Instruct \citep{bai2025qwen3vl} and trained with
% visual-encoder adaptation, grounded SFT, and GRPO, as described in
% Section~\ref{sec:method}. Further details are provided in
% Appendix~\ref{app:dataset_details} and
% Appendix~\ref{app:training_hyperparameters}.

\paragraph{Baselines.} We include four groups of comparisons. (1) Proprietary models: GPT-5.5, Claude Sonnet 5, and Gemini 3.5 Flash. (2) General-purpose models: InternVL2.5-8B and Qwen3-VL-8B, with the latter providing the same-backbone reference before our table-specific training. (3) Table-specialized models: Table-LLaVA~\citep{zheng2024tablellava}, TabPedia~\citep{zhao2024tabpedia}, HIPPO~\citep{wang2026hippo}, and TURBO~\citep{jiang2025turbo};  since TURBO is not publicly released, we report its published results and do not evaluate it on the OOD datasets. (4) Text-based models: Qwen3-8B and Qwen3-8B + RL, which reason over serialized cell contents. The RL variant evaluates whether image-based reasoning remains beneficial when the text-based alternative also receives reasoning post-training. Baseline details are in Appendix~\ref{app:baseline_details}.

\paragraph{Results.}
We first examine whether reasoning over table images improves table question answering and fact verification over text-based and table-specialized methods. As shown in Table~\ref{tab:main_results}, table images alone support strong reasoning: with image-only input, \textsc{Merit} reaches 85.45\% average accuracy, 9.40 points above its Qwen3-VL-8B backbone (more than 11 points on WTQ and TabFact). It also outperforms TURBO, the strongest table-specialized baseline, by 4.36 points and the text-only RL baseline by 2.83 points. Moreover, despite having only 8B parameters, this image-only model surpasses Claude Sonnet 5 and approaches GPT-5.5 and Gemini 3.5 Flash, and our variants obtain the best TABMWP and TAT-QA scores among all methods. Beyond serving as a standalone input, table images also complement serialized text: with the same checkpoint, text-only input obtains 83.42\%, whereas adding the table image raises the average to 86.70\%, the best among open models. This indicates that images supply evidence that serialization loses.

\paragraph{OOD generalization.}
\label{sec:ood_results}
Because SFT and GRPO use only five datasets, strong held-out results could still reflect dataset-specific fitting. We therefore test whether the learned perception and reasoning transfer to unseen datasets with different table structures and visual presentations, including three without serialized tables, which mirror the image-only setting that motivates this work. As shown in Table~\ref{tab:ood_results}, \textsc{Merit} achieves the best average accuracy of 54.90\% on the five unseen datasets, 2.63 points above its Qwen3-VL-8B initialization, with the largest gain on FinMMR (+15.65). In contrast, existing table-specialized models transfer poorly: HIPPO, the strongest among them, averages 25.48\%, less than half of \textsc{Merit}. \textsc{Merit} also exceeds the text-only RL baseline by 12.50 points, indicating that the gains from our training extend beyond the training datasets.
 
% To assess generalization beyond the datasets used for reasoning post-training, we evaluate the final model on five additional datasets without further fine-tuning (Table~\ref{tab:ood_results}). This evaluation examines whether the learned visual and reasoning capabilities transfer to different table structures and presentations. HiTab tests hierarchical-table reasoning, and FinQA tests financial numerical reasoning; both support comparisons with models receiving serialized tables.

% FinMMR, WTB, and MMMU-T cover multimodal financial evidence, naturally occurring table images, and table-related questions in a multidisciplinary multimodal evaluation, respectively. We supply no serialized tables on these datasets, testing whether the model can use visual evidence when a textual table is unavailable. Comparison with the base Qwen3-VL-8B assesses whether our training strengthens this ability beyond the datasets used for SFT and GRPO.

\begin{table*}[t]
    \vspace{-14mm}
\centering
\caption{
Generalization ability on five additional datasets without further fine-tuning.
Scores are accuracy (\%); Avg.\ requires all five scores.
}
\label{tab:ood_results}
\setlength{\tabcolsep}{6pt}
\renewcommand{\arraystretch}{0.95}
\small
\begin{tabular}{lcccccc}
\toprule
\multirow{2}{*}{\textbf{Method}}
& \multicolumn{2}{c}{\textbf{Serialized Table Available}}
& \multicolumn{3}{c}{\textbf{No Serialized Table}}
& \multirow{2}{*}{\textbf{Avg.}} \\
\cmidrule(lr){2-3}
\cmidrule(lr){4-6}
& HiTab & FinQA
& FinMMR & TableVista & MMMU-T
& \\
\midrule
\multicolumn{7}{l}{\textit{General-purpose models}} \\
InternVL2.5-8B
& 31.03 & 24.96 & 25.20 & 35.30 & 22.35 & 27.77 \\
Qwen3-VL-8B
& \underline{67.43} & \textbf{48.55} & 33.02 & \underline{69.54} & \textbf{42.77} & \underline{52.27} \\
Qwen3-8B
& 35.28 & 33.04 & 33.18 & 54.00 & 32.75 & 37.65 \\
\midrule
\multicolumn{7}{l}{\textit{Table-specialized models}} \\
Table-LLaVA-7B
& 7.70 & 12.30 & 6.71 & 19.28 & 5.92 & 10.38 \\
TabPedia
& 9.64 & 2.01 & 0.66 & 10.98 & 1.78 & 5.01 \\
HIPPO
& 35.30 & 30.08 & 16.99 & 33.87 & 11.18 & 25.48 \\
\midrule
\multicolumn{7}{l}{\textit{Text-only RL baseline}} \\
Qwen3-8B + RL
& 39.91 & 45.60 & \underline{36.65} & 57.11 & 32.75 & 42.40 \\
\midrule
\multicolumn{7}{l}{\textit{Our model}} \\
\textbf{Ours}
& \textbf{68.91} & \underline{47.60} & \textbf{48.67} & \textbf{70.72} & \underline{38.60} & \textbf{54.90} \\

\bottomrule
\end{tabular}
\vspace{-5mm}
\end{table*}

\paragraph{Ablation study.}
Finally, we examine why each training stage is needed. \textsc{Merit} stacks three stages, but only SFT and GRPO optimize grounded reasoning on the target tasks, whereas encoder adaptation targets perception. Table~\ref{tab:posttrain_effect} therefore ablates encoder adaptation, SFT, and GRPO under image-only input, enabling each stage alone and in combination while keeping their original order. Encoder adaptation alone leaves end-to-end accuracy nearly unchanged (76.07\% versus 76.05\%), since its auxiliary objectives do not train the complete evidence-to-answer procedure. Its benefit instead emerges after post-training: the full pipeline outperforms SFT + GRPO without adaptation by 0.95 points, with gains on all five datasets. Moreover, the two post-training stages are complementary, as the full pipeline exceeds encoder adaptation followed by GRPO alone by 9.47 points and by SFT alone by 2.11 points. Appendices~\ref{app:encoder_objective_ablation} and \ref{app:reward_component_ablation} further ablate the adaptation objectives and reward components.

\begin{table}[!h]
\vspace{-3mm}
\centering
\caption{Ablation of encoder adaptation and post-training. Scores are accuracy (\%). 
$\Delta$Avg. is relative to the base.}
\label{tab:posttrain_effect}
\small
\setlength{\tabcolsep}{5pt}
\begin{tabular}{lccccccc}
\toprule
\multirow{2}{*}{Variant}
& \multicolumn{3}{c}{Table QA}
& \multicolumn{2}{c}{Fact Verification}
& \multicolumn{2}{c}{Avg.} \\
\cmidrule(lr){2-4}\cmidrule(lr){5-6}\cmidrule(lr){7-8}
& TABMWP & WTQ & TAT-QA & TabFact & InfoTabs & Avg. & $\Delta$Avg. \\
\midrule
\multicolumn{8}{l}{\textit{w/o Encoder adaptation}} \\
Base            & 85.97 & 61.04 & 82.51 & 77.71 & 73.04 & 76.05 & --   \\
+ SFT           & 97.38 & 71.52 & 86.53 & 88.49 & 78.39 & 84.46 & +8.41 \\
+ GRPO          & 93.68 & 60.52 & 81.74 & 77.93 & 73.13 & 77.40 & +1.35 \\
+ SFT + GRPO    & 97.70 & 71.52 & 86.14 & 89.38 & 77.78 & 84.50 & +8.45 \\
\midrule
\multicolumn{8}{l}{\textit{w/ Encoder adaptation}} \\
Encoder only    & 86.44 & 61.35 & 82.38 & 77.46 & 72.70 & 76.07 & +0.01   \\
+ SFT           & 97.09 & 68.85 & 85.10 & 87.48 & 78.17 & 83.34 & +7.28 \\
+ GRPO          & 91.02 & 61.46 & 79.53 & 78.19 & 69.67 & 75.98 & -0.08 \\
+ SFT + GRPO 
                & 98.13 & 72.86 & 87.56 & 90.08 & 78.63 & 85.45 & +9.40 \\
\bottomrule
\end{tabular}
\vspace{-7mm}
\end{table}

\section{Conclusion}
We presented \textsc{Merit}, which reasons over table images as primary evidence, with serialized text as optional input. Guided by a diagnostic that identifies visual perception as a bottleneck, it adapts the visual encoder with table-centric objectives and then applies evidence-grounded SFT and RL with decomposed verifiable rewards. Image-only \textsc{Merit} outperforms its backbone, table-specialized models, and a text-only RL baseline, approaches proprietary models, and transfers to unseen datasets; adding serialized text improves it further. Ablations show that encoder adaptation pays off after post-training and that SFT and GRPO are complementary.

%\section*{Limitations}

% The workspace bibliography is refs.bib. The ICLR template should also provide
% iclr2027_conference.bst.
\bibliography{refs}
\bibliographystyle{iclr2027_conference}

\clearpage
\begin{center}
{\Large\bfseries Appendix}
\end{center}
\vspace{2mm}
\appendix

\section{Training and Data Details}
\label{app:training_data_details}

\subsection{Dataset Details}
\label{app:dataset_details}

\paragraph{Visual-adaptation data.}
For Stage 1, we construct a structured-table corpus for adapting the visual encoder through table reconstruction, cell localization, and color recognition. Data construction and statistics are provided in Appendix~\ref{app:visual_adaptation_data}.

\paragraph{Post-training data.}
For SFT and GRPO, we use training examples from TABMWP~\citep{lu2023tabmwp}, WikiTableQuestions (WTQ)~\citep{pasupat2015wtq}, TAT-QA~\citep{zhu2021tatqa}, TabFact~\citep{chen2020tabfact}, and InfoTabs~\citep{gupta2020infotabs}. TABMWP contains mathematical word problems, WTQ requires compositional question answering, and TAT-QA combines tables with financial text; TabFact and InfoTabs are fact-verification datasets with binary and three-way labels, respectively.

\paragraph{Evaluation data.}
In addition to the datasets used for post-training, we include HiTab~\citep{cheng2022hitab}, FinQA~\citep{chen2021finqa}, FinMMR~\citep{tang2025finmmr}, TableVista~\citep{yang2026tablevista}, and MMMU-T~\citep{yue2024mmmu} to evaluate the model's generalization to unseen table-reasoning tasks and domains.

\subsection{Visual-Adaptation Data Details}
\label{app:visual_adaptation_data}

All Stage-1 examples are built from MMTab~\citep{zheng2024tablellava}. For table reconstruction, the model reproduces the serialized table from its image. For cell localization, the model answers position-based queries, returning the content at a given row and column or the position of a given value. For color recognition, we assign random background colors to selected cells of MMTab tables, re-render the table images, and pose questions whose answers depend on these colors, such as the value in a cell of a given color or the color of a given cell. Because colors are sampled independently of cell values, these questions cannot be answered from the textual content alone. We additionally include table question-answering examples from MMTab as an auxiliary task. All tasks are mixed without task-specific weighting; Table~\ref{tab:visual_adaptation_statistics} summarizes the composition.

\begin{table}[!h]
    \centering
    \small
    \setlength{\tabcolsep}{6pt}
    \begin{tabular}{lcc}
        \toprule
        Task & Data source & Training examples \\
        \midrule
        Table reconstruction & MMTab             & 65,068 \\
        Cell localization    & MMTab             & 34,349 \\
        Color recognition    & MMTab (recolored) & 34,349 \\
        Table QA (auxiliary) & MMTab             & 50,000 \\
        \midrule
        Total                & --                & 183,766 \\
        \bottomrule
    \end{tabular}
    \caption{
    Statistics of the visual-adaptation data used in Stage 1.
    }
    \label{tab:visual_adaptation_statistics}
\end{table}

The instruction templates used for the visual-adaptation tasks are summarized in Table~\ref{tab:visual_adaptation_templates}.

\begin{table}[!h]
    \centering
    \small
    \setlength{\tabcolsep}{6pt}
    \begin{tabular}{lp{0.68\linewidth}}
        \toprule
        Task & Instruction template \\
        \midrule
        Table reconstruction &
        Reconstruct the structure and content of the table from the given table image. \\
        Cell localization &
        Locate the logical cell corresponding to the specified target in the given table image. \\
        Color recognition &
        Identify the visual color attribute of the specified cell in the given table image. \\
        Table QA (auxiliary) &
        The original MMTab question for the given table image. \\
        \bottomrule
    \end{tabular}
    \caption{
    Instruction templates for the visual-adaptation tasks.
    }
    \label{tab:visual_adaptation_templates}
\end{table}

\subsection{SFT Data Composition}
\label{app:sft_data_composition}

Table~\ref{tab:sft_data_statistics} reports the statistics of the SFT data before and after trajectory verification. For each dataset, we report the number of generated trajectories and the number that pass all three verification checks and enter the SFT set.

\begin{table}[!h]
    \centering
    \small
    \setlength{\tabcolsep}{6pt}
    \begin{tabular}{lcc}
        \toprule
        Dataset & Generated & Verified \\
        \midrule
        TABMWP   & 2,000 & 1,540 \\
        WTQ      & 2,000 & 1,587 \\
        TAT-QA   & 2,000 & 1,660 \\
        TabFact  & 2,000 & 1,382 \\
        InfoTabs & 2,000 & 1,856 \\
        \bottomrule
    \end{tabular}
    \caption{
    Numbers of generated and verified SFT trajectories per dataset.
    }
    \label{tab:sft_data_statistics}
\end{table}

Rejected trajectories primarily contain cell references inconsistent with the source table, invalid intermediate calculations, or final answers inconsistent with the annotated targets. We also remove duplicate examples and exclude samples that overlap with the evaluation splits.

\subsection{Generation and Verification Prompts}
\label{app:generation_verification_prompts}

For reproducibility, we provide the complete prompts used by the trajectory generator and the independent multimodal verifier.

\paragraph{Trajectory-generation prompt.}
\begin{quote}
\small
You are constructing grounded chain-of-thought training data for table reasoning. You are given a table image, the structured table with cell metadata (logical row, column, value, and visual attributes such as background color), a question or statement, and its annotated answer. Write a reasoning trajectory that a model could produce from the table image alone and that reaches the annotated answer.

Requirements:
\begin{itemize}\setlength{\itemsep}{0pt}
\item Identify the headers and cells relevant to the question. Cite every cell you use exactly as \texttt{"value" [cell: (tN,rN,cN)]}, where \texttt{r} and \texttt{c} are the logical row and column of the cell in the structured table and \texttt{value} is its normalized value.
\item Check each cited value against the corresponding cell before using it in a lookup, comparison, calculation, or factual inference.
\item If a visual condition, such as a cell color or highlighting, determines which cells are relevant, state it explicitly and cite only the cells that satisfy it.
\item For calculation questions, write every arithmetic step as an equality such as \mbox{\texttt{12 + 7 = 19}}, using only cited values or earlier intermediate results as operands.
\item Do not mention the structured table, the cell metadata, or the annotated answer; the reasoning must read as if derived from the image.
\end{itemize}
Output format:\\
\texttt{<think>} reasoning with cited cells and calculations \texttt{</think>}\\
\texttt{<answer>} final answer or verification label \texttt{</answer>}

Table image: \{image\}\\
Structured table with cell metadata: \{table\}\\
Question: \{question\}\\
Annotated answer: \{answer\}
\end{quote}

\paragraph{Trajectory-verification prompt.}
\begin{quote}
\small
You are verifying a grounded table-reasoning trajectory. You are given the table image, the structured table with cell metadata, the question or statement, the annotated answer, and the trajectory. Check the following three conditions:
\begin{enumerate}\setlength{\itemsep}{0pt}
\item \textbf{Evidence}: every cited cell \texttt{[cell: (tN,rN,cN)]} exists in the structured table, its cited value matches the cell value, and the cited cells are sufficient and relevant for the question, including any visual condition such as color.
\item \textbf{Derivation}: every lookup, comparison, and calculation step is valid, and every operand comes from a cited cell or an earlier intermediate result.
\item \textbf{Answer}: the final answer in \texttt{<answer>}\ldots\texttt{</answer>} matches the annotated answer after normalization.
\end{enumerate}
If all three conditions hold, reply exactly: CORRECT\\
Otherwise, reply exactly: WRONG\\
Do not explain.

Table image: \{image\}\\
Structured table with cell metadata: \{table\}\\
Question: \{question\}\\
Annotated answer: \{answer\}\\
Trajectory: \{trajectory\}
\end{quote}

The generated trajectories identify the relevant headers and cells, cite each cell by its logical coordinates and normalized value, state the necessary intermediate calculations, and end with the final answer. The verifier checks the cited evidence against the source inputs, the validity of the derivation, and the correctness of the final answer; only trajectories passing all three checks are retained.

\subsection{RL Data Composition}
\label{app:rl_data_composition}

Table~\ref{tab:rl_data_statistics} reports the number of GRPO training prompts drawn from each of the five datasets.

\begin{table}[!h]
    \centering
    \small
    \setlength{\tabcolsep}{6pt}
    \begin{tabular}{llc}
        \toprule
        Dataset & Task & Prompts \\
        \midrule
        TABMWP   & Table QA          & 2,000 \\
        WTQ      & Table QA          & 2,000 \\
        TAT-QA   & Table QA          & 2,000 \\
        TabFact  & Fact verification & 2,000 \\
        InfoTabs & Fact verification & 2,000 \\
        \bottomrule
    \end{tabular}
    \caption{
    Composition of the data used for GRPO training. 
    }
    \label{tab:rl_data_statistics}
\end{table}

For each input question, we sample \(G=8\) candidate responses from the current policy. The reward formulation is described in Section~\ref{sec:table_rl}, and its weights and parsing rules are given in Appendix~\ref{app:reward_parsing_details}.

\subsection{Reward and Parsing Details}
\label{app:reward_parsing_details}

We set \(\lambda_{\mathrm{ans}}=1\) and \(\lambda_{\mathrm{evi}}=\lambda_{\mathrm{calc}}=\lambda_{\mathrm{fmt}}=0.5\) in Eq.~\eqref{eq:total_reward}. The final answer is extracted from the \answeropen{} field and compared with the annotation after normalization: strings are lowercased and stripped of whitespace, punctuation, currency symbols, and thousands separators; numeric answers are compared exactly after parsing, with percentages converted to fractions; multi-value answers are compared as sets; and fact-verification labels are mapped to canonical classes. Evidence cells are parsed from the \texttt{[cell: (t, r, c)]} annotations and matched against the table by both logical coordinate and value. Calculations are parsed as arithmetic expressions and executed, and their results are compared with the verified reference calculation; each operand must come from a cited cell or an earlier intermediate result.

\subsection{Training Hyperparameters}
\label{app:training_hyperparameters}

All experiments use Qwen3-VL-8B-Instruct as the base model. The principal hyperparameters for the three training stages are summarized in Table~\ref{tab:training_hyperparameters}.

\begin{table}[!h]
    \centering
    \small
    \setlength{\tabcolsep}{5pt}
    \begin{tabular}{lccc}
        \toprule
        Hyperparameter
        & Visual adaptation
        & SFT
        & GRPO \\
        \midrule
        Learning rate
        & 5e-6 & 1e-5 & 2e-6 \\
        Batch size
        & 24 & 128 & 64 prompts \\
        Gradient accumulation steps
        & 3 & 16 & -- \\
        Training epochs / steps
        & 3 / 6,782 & 2 / 126 & 5 / 780 \\
        Maximum sequence length
        & 4,608 & 3,328 & 18,432 + 1,024 \\
        Optimizer
        & AdamW & AdamW & AdamW \\
        Learning-rate scheduler
        & Cosine & Cosine & Constant \\
        Warmup ratio
        & 0.03 & 0.03 & 0 \\
        Weight decay
        & 0 & 0 & 0.01 \\
        Numerical precision
        & BF16 & BF16 & BF16 \\
        Trainable modules
        & Visual encoder
        & Language decoder
        & Language decoder \\
        Responses per group \(G\)
        & -- & -- & 8 \\
        Clipping coefficient
        & -- & -- & 0.2 \\
        KL coefficient
        & -- & -- & 0 \\
        \bottomrule
    \end{tabular}
    \caption{
    Training hyperparameters for table-aware visual adaptation, supervised fine-tuning, and GRPO.
    }
    \label{tab:training_hyperparameters}
\end{table}

\section{Evaluation and Baseline Details}
\label{app:baseline_details}

\subsection{Evaluation Protocol and Scoring}
\label{app:evaluation_details}

The five post-training datasets---TABMWP, WTQ, TAT-QA, TabFact, and InfoTabs---are used for supervised fine-tuning and GRPO. We evaluate the resulting model on the corresponding held-out splits, which constitute the in-distribution (IID) setting. The additional datasets---HiTab, FinQA, FinMMR, TableVista, and MMMU-T---are used only for evaluation and constitute the out-of-domain (OOD) setting. No additional fine-tuning is performed on the OOD datasets.

Unless otherwise specified, the model receives the table image, the question or statement, and accompanying passages when available; serialized tables are supplied only in the specified input comparisons. On the OOD datasets, only HiTab and FinQA provide serialized tables; for FinMMR, TableVista, and MMMU-T, we convert the table images into Markdown with MinerU~2.5~\citep{niu2026mineru25}. Text-based models receive this textual table, and visual models receive both the table image and the textual table.

Answers are extracted and evaluated against the ground truth, with ambiguous cases reviewed by GPT-5.5. For Table QA, we report accuracy and the unweighted average of the per-dataset accuracies. For table fact verification, we report accuracy under the corresponding binary or three-way label setting.

In Table~\ref{tab:main_results}, $\dagger$ marks scores taken from published reports under their original evaluation protocols. The remaining scores are obtained using our evaluation pipeline, detailed in Appendix~\ref{app:evaluation_details}.

\subsection{Table-Structure Understanding Evaluation}
\label{app:structure_eval}

Table~\ref{tab:heldin} and Table~\ref{tab:encoder_task_ablation} evaluate the visual encoder on six held-in table-structure understanding tasks from MMTab~\citep{zheng2024tablellava}. We follow the task definitions, metrics, and official evaluation scripts of Table-LLaVA~\citep{zheng2024tablellava}. \textbf{Table Size Detection (TSD)} asks for the numbers of rows and columns, and we report row and column accuracy. \textbf{Table Cell Extraction (TCE)} asks for the content of the cell at a given row and column, and \textbf{Table Cell Locating (TCL)} asks for the position of a given cell; both report accuracy. \textbf{Merged Cell Detection (MCD)} asks the model to identify merged cells and reports F1. \textbf{Row and Column Extraction (RCE)} asks for all cells in a given row or column and reports row and column F1. \textbf{Table Recognition (TR)} converts the table image into HTML, Markdown, or LaTeX and reports Tree-Edit-Distance-based Similarity (TEDS) for each format, with Markdown and LaTeX outputs converted to HTML before scoring. For tasks with multiple submetrics, we first average the submetrics; Avg.\ denotes the macro-average over the six tasks.

\subsection{General-Purpose Vision-Language Models}
\label{app:general_baselines}

We evaluate InternVL2.5-8B and Qwen3-VL-8B-Instruct, abbreviated as Qwen3-VL-8B in the result tables. These baselines are evaluated without our table-specific training. Qwen3-VL-8B-Instruct provides the initialization of \textsc{Merit} and serves as a reference for measuring the overall effect of visual-encoder adaptation, grounded SFT, and GRPO.

In the in-distribution comparisons, these models receive table images, questions or verification statements, and accompanying passages when available, without serialized table contents. On the OOD datasets, they follow the input setting in Appendix~\ref{app:evaluation_details}.

\subsection{Table-Specialized Models}
\label{app:table_baselines}

We include Table-LLaVA-7B, TabPedia-7B, HIPPO-8B, and TURBO-8B. Their per-dataset scores in Table~\ref{tab:main_results} are taken from Table~1 of \citet{jiang2025turbo}. We retain these scores and recompute their unweighted averages over TABMWP, WTQ, TAT-QA, TabFact, and InfoTabs. For the OOD evaluation in Table~\ref{tab:ood_results}, we evaluate the released Table-LLaVA, TabPedia, and HIPPO checkpoints with our pipeline under the input setting in Appendix~\ref{app:evaluation_details}; TURBO is not publicly released and is therefore not evaluated on these datasets.

The reported input configurations differ across methods. HIPPO uses both table images and structured tables; the reported scores do not correspond to its image-only ``HIPPO w/o ST'' variant. TURBO receives table images at inference, while structured tables are used during training-data construction.

For the table-perception evaluation in Table~\ref{tab:heldin}, we additionally include the 7B and 13B variants of Table-LLaVA~\citep{zheng2024tablellava}. These comparisons assess table structure understanding separately from the downstream reasoning evaluation in Table~\ref{tab:main_results}.

\subsection{Proprietary Models}
\label{app:proprietary_baselines}
We evaluate GPT-5.5~\citep{openai2026gpt55}, Claude Sonnet 5, and Gemini 3.5 Flash through their APIs. Each model receives the table image and the original question, and responses are generated with temperature 0. Their outputs are scored with the same protocol as the other models (Appendix~\ref{app:evaluation_details}).

\subsection{Text-Based Models}
\label{app:text_baselines}

We evaluate Qwen3-8B and Qwen3-8B + RL using serialized table contents without direct access to table images. Their inputs also include questions or verification statements and accompanying passages when available. Qwen3-8B + RL is trained with the same SFT and GRPO settings as \textsc{Merit}, with serialized tables in place of table images, providing a textual baseline that also undergoes reasoning post-training.

These baselines differ from Ours (Text only), which uses \textsc{Merit}'s final checkpoint without table images. Ours (Text only), Ours (Image only), and Ours (Image + Text) share this checkpoint and differ only in the supplied table representation, allowing input modalities to be compared with the model fixed.

\section{Additional Analyses and Case Studies}
\label{app:more_experiments}

\subsection{Error Analysis}
\label{app:encoder_badcase}
To characterize where models fail and how training changes the error profile, we analyze errors on a stratified IID subset (299 examples per dataset, 1,495 in total) with image-only input, together with a color probe of 486 questions on MMTab tables with randomly colored cells. For the IID subset, an LLM classifier compares each wrong response with the ground-truth table and assigns its primary cause to one of five categories: \emph{recognition} (misread or missed cell content), \emph{locating} (values taken from the wrong row, column, or header), \emph{color} (a visual attribute ignored or misidentified), \emph{reasoning} (correct evidence but a wrong calculation, comparison, or inference), and \emph{comprehension} (a misunderstood question). Errors on the color probe are attributed by rules against the known coloring. Cases that are not model errors, namely format-only mismatches, correct answers missed by the scorer, and questionable gold labels, are excluded, as are responses whose classification could not be parsed (about 2\%).

\begin{figure}[!h]
    \centering
    \begin{subfigure}{\linewidth}
        \centering
        \includegraphics[width=0.9\linewidth]{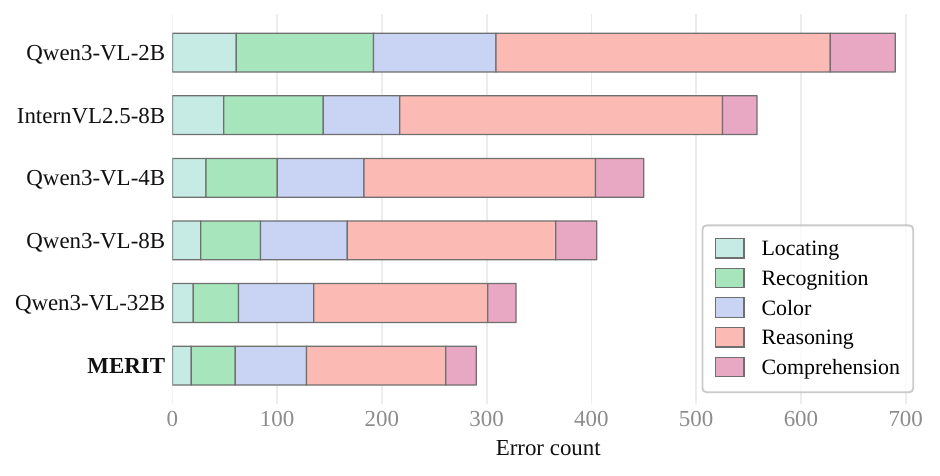}
        \caption{Error counts by category on the IID subset and the color probe.}
    \end{subfigure}

    \vspace{2mm}
    \begin{subfigure}{\linewidth}
        \centering
        \includegraphics[width=0.75\linewidth]{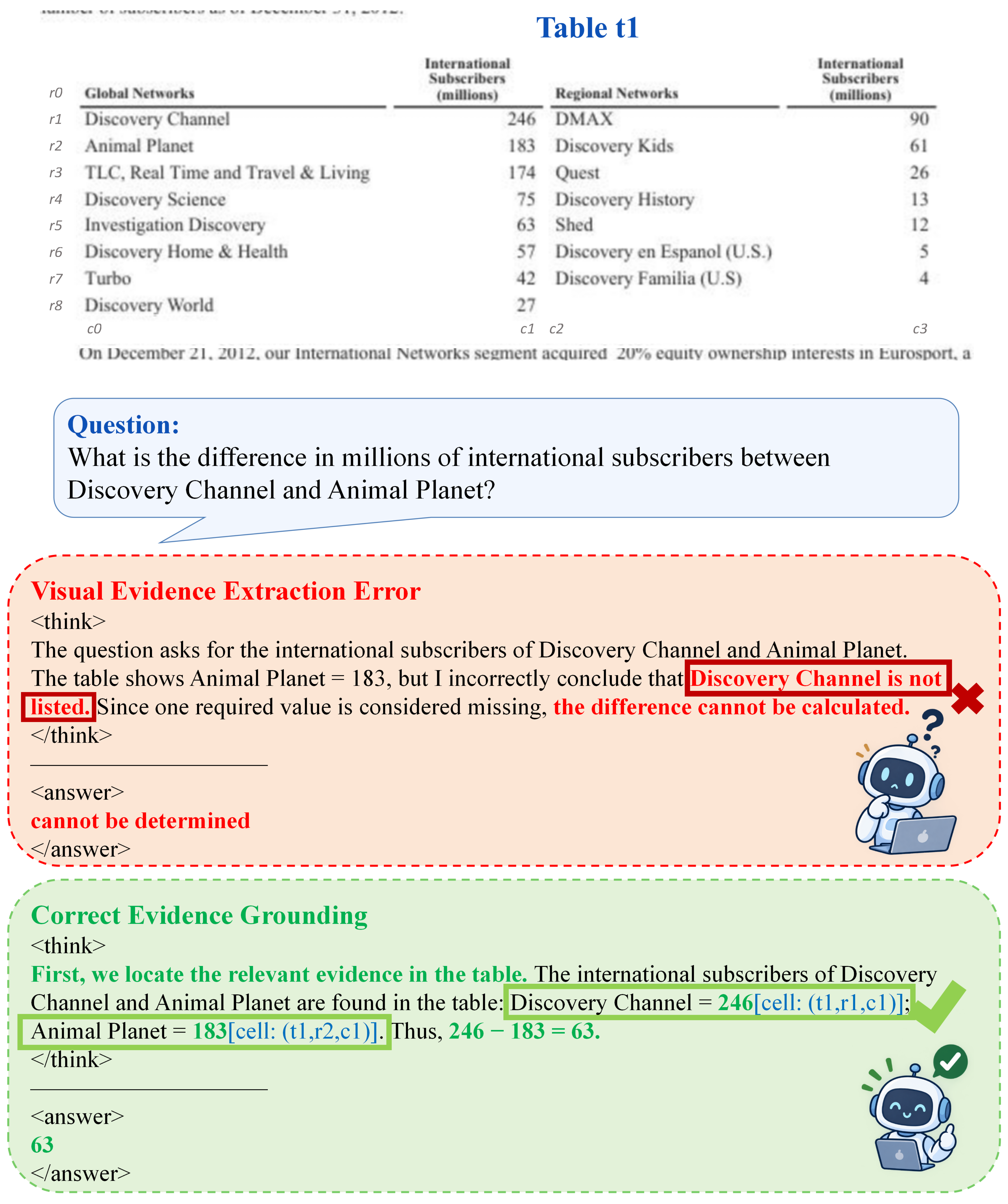}
        \caption{A visual evidence extraction error of the unadapted encoder and the corresponding correct grounding.}
    \end{subfigure}
    \caption{Error analysis. (a) Error distribution of general-purpose VLMs and \textsc{Merit}. (b) A representative perception failure.}
    \label{fig:error_analysis}
    \label{fig:encoder_badcase}
\end{figure}

As shown in Figure~\ref{fig:error_analysis}(a), perception errors (recognition, locating, and color) account for about two fifths of the errors of Qwen3-VL-8B (167 of 405). \textsc{Merit} reduces errors in every category, including recognition (57$\rightarrow$42), locating (27$\rightarrow$18), color (83$\rightarrow$68), and reasoning (199$\rightarrow$133), for a total of 290 errors, fewer than Qwen3-VL-32B (328). Color remains a major error source for all models, mostly from counting colored cells. Figure~\ref{fig:error_analysis}(b) shows a representative perception failure of the unadapted encoder: it misses the Discovery Channel row and concludes that the required value is not listed, whereas correct grounding locates both cells, $(t1,r1,c1)$ and $(t1,r2,c1)$, and computes $246-183=63$.

\subsection{Visual Encoder Objective Ablation}
\label{app:encoder_objective_ablation}

We examine the three objectives used to adapt the encoder: \textsc{Recon} for content and structure, \textsc{Locate} for associations between values and grid positions, and \textsc{Color} for visual-attribute grounding. Table~\ref{tab:encoder_task_ablation} compares individual objectives, pairs, and the full objective before SFT and GRPO. Removing each objective from Full evaluates its contribution. Structure, localization, and color-conditioned metrics identify the perceptual abilities affected, while QA accuracy measures their downstream benefit before reasoning post-training.

The full objective achieves the highest average score of 53.51,
compared with 46.89 for the base model. Among the individual objectives,
\textsc{Recon} and \textsc{Locate} reach average scores of 48.82 and
47.71, respectively, whereas \textsc{Color} alone scores 40.25.
Combining \textsc{Recon} and \textsc{Locate} yields 52.73. The
leave-one-objective-out comparisons show that removing \textsc{Locate}
causes the largest drop from the full objective (11.56 points), followed
by removing \textsc{Recon} (7.04 points); removing \textsc{Color} reduces
the average by 0.78 points. These results suggest that reconstruction
and localization are the main contributors to the reported aggregate
perception score, while \textsc{Color} provides a smaller incremental
gain.

\begin{table}[!h]
\centering
\caption{
Encoder-objective ablation before SFT and GRPO.
Tasks and metrics are described in Appendix~\ref{app:structure_eval}.
Avg.\ first averages the submetrics within each benchmark and then
computes the macro-average over the six held-in benchmarks.
Scores are accuracy, F1, or TEDS, depending on the benchmark.
}
\label{tab:encoder_task_ablation}
\setlength{\tabcolsep}{5pt}
\renewcommand{\arraystretch}{1.12}
\resizebox{\textwidth}{!}{%
\begin{tabular}{lccccccccccc}
\toprule
\multirow{2}{*}{\textbf{Encoder Objective}}
& \multicolumn{2}{c}{\textbf{TSD}}
& \textbf{TCE}
& \textbf{TCL}
& \textbf{MCD}
& \multicolumn{2}{c}{\textbf{RCE}}
& \multicolumn{3}{c}{\textbf{TR}}
& \multirow{2}{*}{\textbf{Avg.}} \\
\cmidrule(lr){2-3}
\cmidrule(lr){4-4}
\cmidrule(lr){5-5}
\cmidrule(lr){6-6}
\cmidrule(lr){7-8}
\cmidrule(lr){9-11}
& Row & Col.
& Acc.
& Acc.
& F1
& Row & Col.
& HTML & Markdown & LaTeX
& \\
\midrule
Base
& 37.80 & 72.40
& 37.46 & 38.63 & 30.99
& 41.19 & 68.02
& 59.26 & 88.19 & 46.25
& 46.89 \\
Recon only
& 34.70 & 70.60
& 40.06 & 44.79 & 21.44
& 45.41 & 71.06
& 69.74 & 91.33 & 66.23
& 48.82 \\
Locate only
& 26.60 & 46.30
& 43.07 & \textbf{57.57} & 24.51
& 46.56 & 70.09
& 58.08 & 88.49 & 52.40
& 47.71 \\
Color only
& 35.50 & 52.60
& 30.34 & 38.39 & 15.47
& 34.67 & 61.36
& 57.70 & 89.33 & 48.75
& 40.25 \\
Recon + Locate
& 38.80 & 71.20
& \textbf{46.67} & 53.79 & 28.71
& 48.84 & \textbf{76.89}
& 63.28 & 89.75 & 54.96
& 52.73 \\
Recon + Color
& 37.50 & 62.30
& 29.20 & 36.86 & 22.11
& 38.72 & 65.98
& 62.58 & 86.01 & 35.16
& 41.95 \\
Locate + Color
& 34.50 & 54.50
& 37.38 & 49.90 & 23.45
& 43.79 & 68.96
& 58.86 & 88.83 & 53.92
& 46.47 \\
\textbf{Full (Ours)}
& \textbf{45.20} & \textbf{73.20}
& 45.56 & 48.75 & \textbf{35.91}
& \textbf{51.40} & 75.35
& 62.12 & \textbf{89.96} & 52.76
& \textbf{53.51} \\
\bottomrule
\end{tabular}%
}
\end{table}

\subsection{Reward Component Ablation}
\label{app:reward_component_ablation}

Our RL design provides feedback on evidence selection and numerical
calculation in addition to the final answer. Starting from the same
encoder-adapted SFT checkpoint, we compare answer-only supervision with
variants that additionally introduce evidence or calculation rewards.
The Full variant combines all reward components, while keeping the
remaining RL settings fixed (Table~\ref{tab:reward_variant}).

Evidence F1 measures the quality of selected evidence, and calculation accuracy measures computation on arithmetic examples. Final-answer accuracy evaluates whether feedback on these intermediate steps benefits task completion. These measurements connect the reward design to the retrieval and computational errors it is intended to address.

Answer-only supervision achieves an average accuracy of 85.01\%.
Adding the evidence reward raises it slightly to 85.11\% (+0.10 points),
whereas adding the calculation reward yields 85.39\% (+0.38 points).
The Full variant achieves the highest average accuracy, 85.45\%, a
0.44-point gain over answer-only supervision and a 0.06-point gain over
the calculation-reward variant. The gains are not uniform across
datasets: for example, the evidence-reward variant improves WTQ but
scores lower on TAT-QA and InfoTabs, while the calculation-reward
variant improves TAT-QA but slightly lowers TabFact accuracy. Overall,
the results show modest improvements in average final-answer accuracy
from adding intermediate rewards, with the largest individual gain
coming from calculation feedback.

\begin{table}[!h]
\centering
\caption{
Effect of different reward variants with the final checkpoint fixed.
Questions, table inputs, and decoding settings are unchanged.
The three QA datasets are TABMWP, WTQ, and TAT-QA, while the two
fact verification datasets are TabFact and InfoTabs.
Avg.\ averages all five datasets. Scores are accuracy (\%).
}
\label{tab:reward_variant}
\setlength{\tabcolsep}{5pt}
\renewcommand{\arraystretch}{1.12}
\resizebox{0.78\textwidth}{!}{%
\begin{tabular}{lcccccc}
\toprule
\multirow{2}{*}{\textbf{Reward Variant}}
& \multicolumn{3}{c}{\textbf{Table Question Answering}}
& \multicolumn{2}{c}{\textbf{Fact Verification}}
& \multirow{2}{*}{\textbf{Avg.}} \\
\cmidrule(lr){2-4}
\cmidrule(lr){5-6}
& TABMWP & WTQ & TAT-QA
& TabFact & InfoTabs & \\
\midrule
Answer only
& 97.94 & 71.29 & 87.18 & 89.73 & 78.93 & 85.01 \\
Answer + Evidence
& \textbf{98.13} & 72.31 & 87.05 & 89.80 & 78.26 & 85.11 \\
Answer + Calculation
& 98.04 & 72.03 & \textbf{88.08}
& 89.48 & \textbf{79.31} & 85.39 \\
\textbf{Full}
& \textbf{98.13} & \textbf{72.86} & 87.56
& \textbf{90.08} & 78.63 & \textbf{85.45} \\
\bottomrule
\end{tabular}%
}
\end{table}

\subsection{Additional Image--Text Comparison Cases}
\label{app:image_text_cases}

\begin{figure}[!h]
    \centering
    \includegraphics[
        width=0.94\textwidth,
        height=0.42\textheight,
        keepaspectratio]{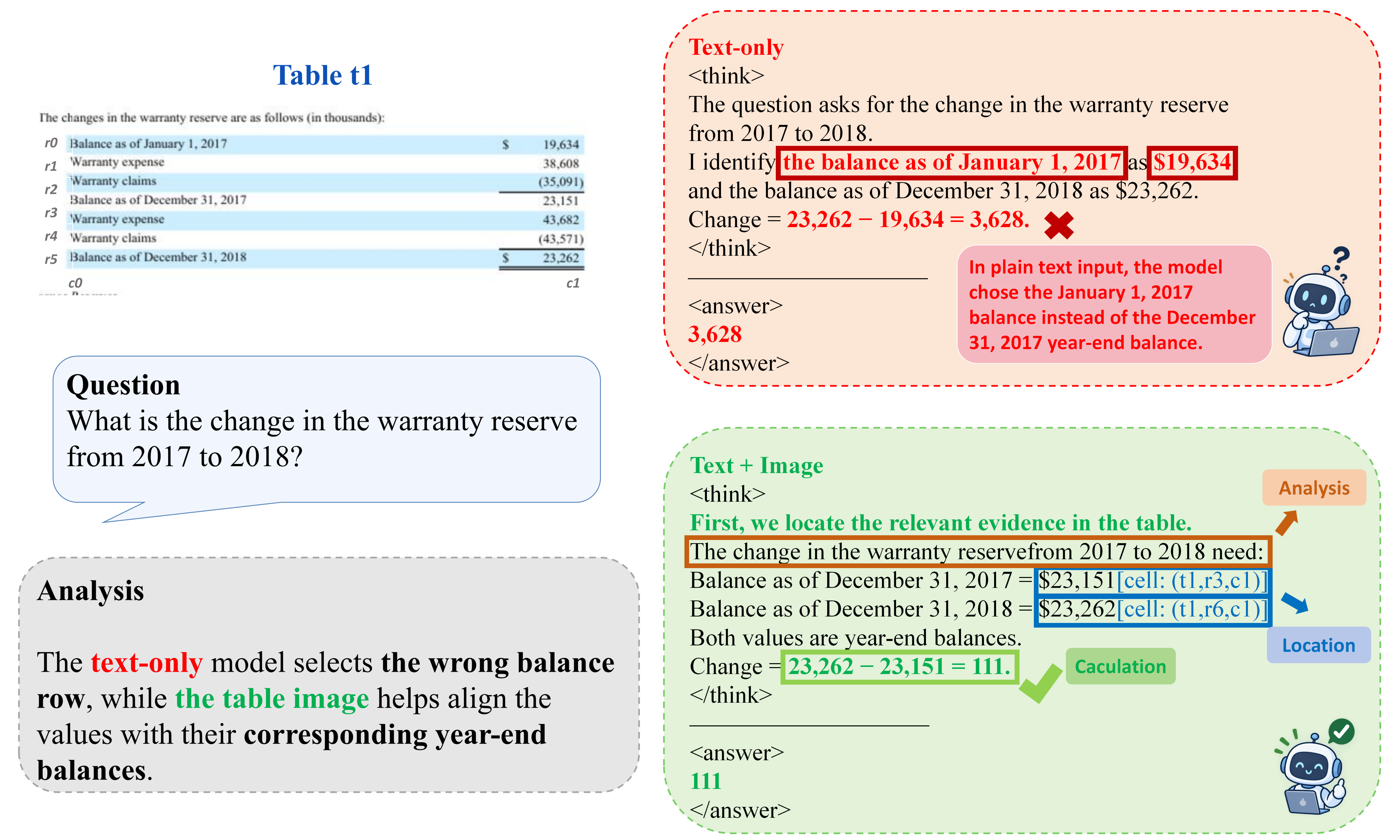}
    \caption{
        Image--text comparison case 1, illustrating temporal row alignment.
    }
    \label{fig:image_matters_case_1}
\end{figure}

\begin{figure}[!h]
    \centering
    \includegraphics[
        width=0.94\textwidth,
        height=0.42\textheight,
        keepaspectratio]{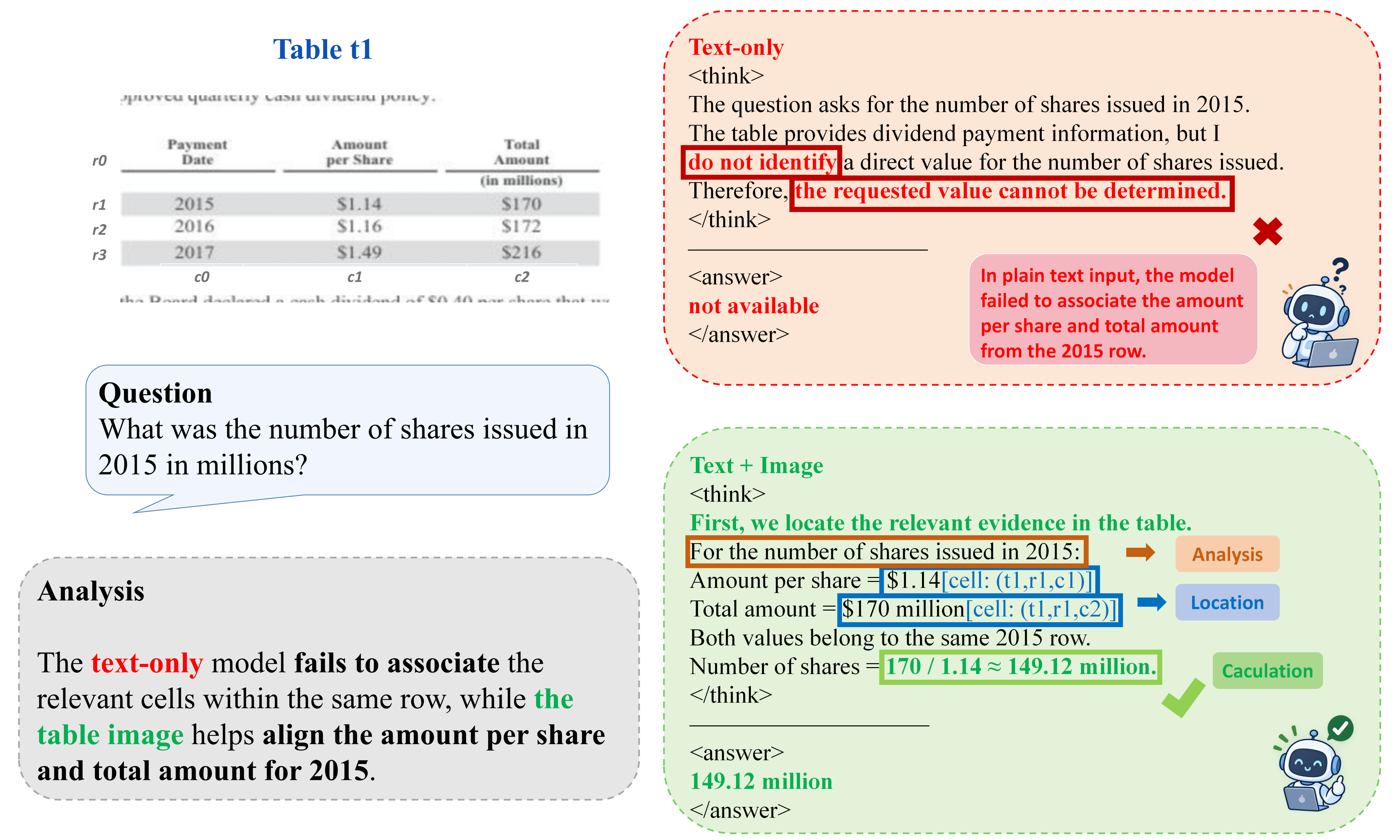}
    \caption{
        Image--text comparison case 2, illustrating within-row association.
    }
    \label{fig:image_matters_case_2}
\end{figure}

Figures~\ref{fig:image_matters_case_1} and~\ref{fig:image_matters_case_2}
present additional qualitative comparisons to illustrate how
table images complement serialized table text.
These cases highlight two types of structural errors under text-only
input: temporal row alignment and within-row association.

\subsection{More Grounded Reasoning Cases}
\label{app:more_grounded_cases}

\begin{figure}[!h]
    \centering
    \includegraphics[
        width=\linewidth,
        trim={8pt 20pt 8pt 6pt},
        clip,
        height=0.42\textheight,
        keepaspectratio]{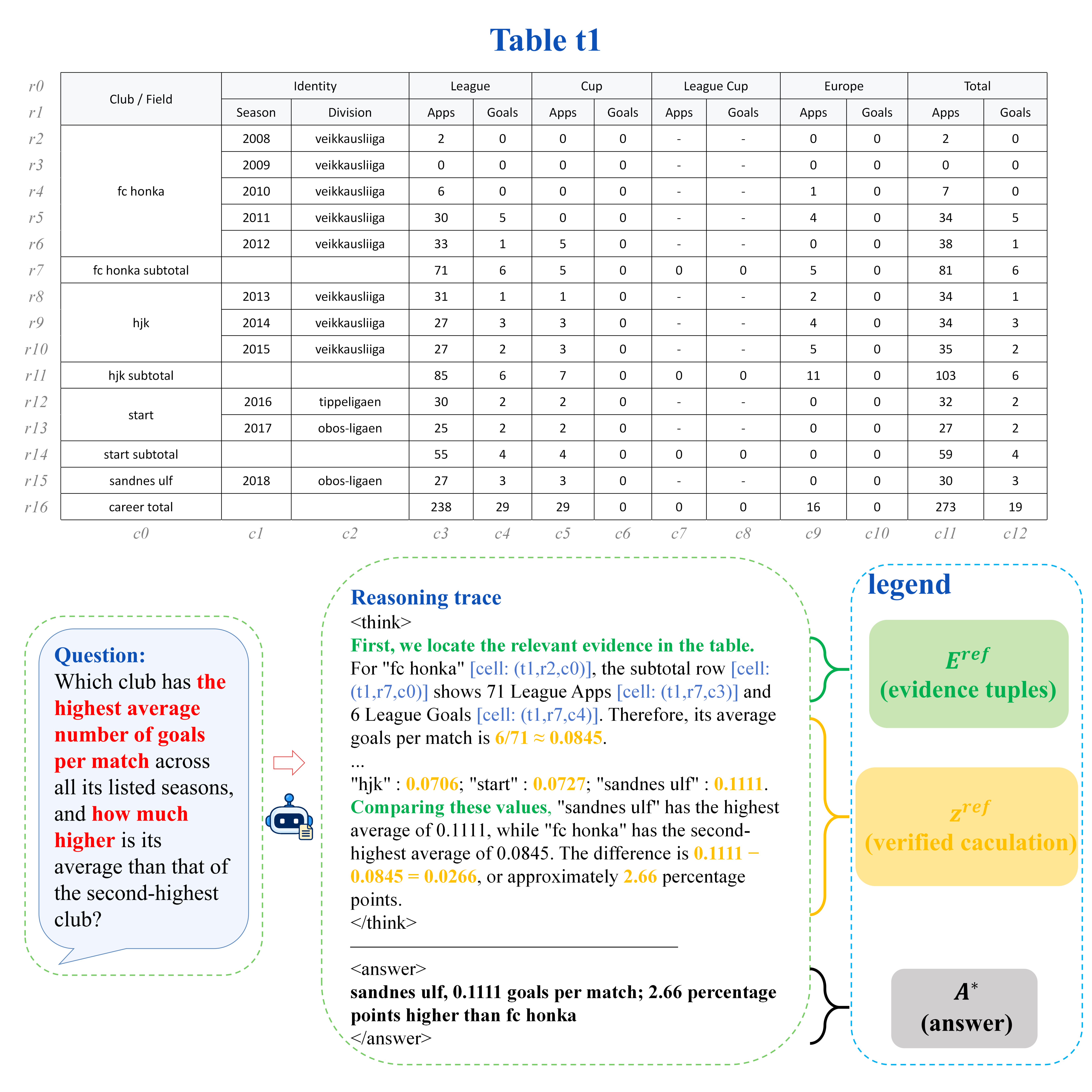}
    \vspace{-1.5mm}
    \caption{
        Grounded reasoning case 1, showing the table, question,
        cited cells, reasoning process, and final answer.
    }
    \label{fig:more_cot_case_1}
    \vspace{-1.5mm}
\end{figure}

\begin{figure}[!h]
    \centering
    \includegraphics[
        width=\linewidth,
        trim={8pt 20pt 8pt 6pt},
        clip,
        height=0.42\textheight,
        keepaspectratio]{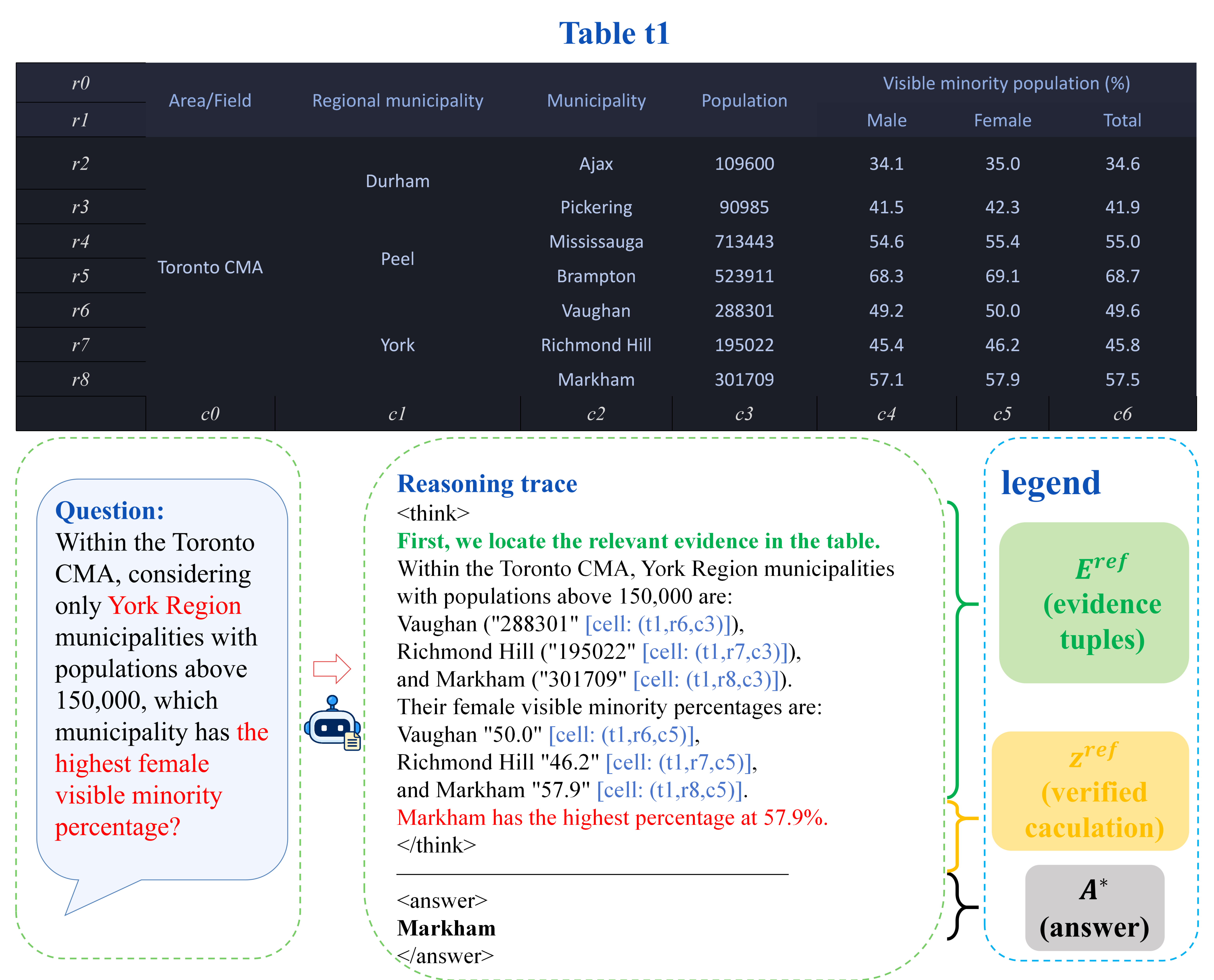}
    \vspace{-1.5mm}
    \caption{
        Grounded reasoning case 2, showing the table, question,
        cited cells, reasoning process, and final answer.
    }
    \label{fig:more_cot_case_2}
    \vspace{-1.5mm}
\end{figure}

We present additional grounded reasoning cases in
Figures~\ref{fig:more_cot_case_1} and~\ref{fig:more_cot_case_2}.
These examples demonstrate how our trajectory format explicitly connects
logical cell coordinates to intermediate calculations and final answers,
so that evidence selection, computation, and the final answer can each be
checked against the source table.

\end{document}